\documentclass[11pt,letterpaper]{article}
\usepackage[T1]{fontenc}
\usepackage[utf8]{inputenc}
\usepackage{lmodern}
\usepackage[margin=1in]{geometry}
\usepackage{amsmath,amssymb,amsthm,mathtools,bm}
\usepackage{graphicx,float}
\usepackage{array,booktabs}
\usepackage{enumitem}
\usepackage[numbers,sort&compress]{natbib}
\usepackage[hyphens]{url}
\usepackage[hidelinks]{hyperref}
\hypersetup{
  pdftitle={Two-Layer Random ReLU Networks Can Separate Disjoint Classes},
  pdfauthor={Yichen Liu and Jianbin Yang}
}
\newcommand{\DOI}[1]{DOI~\href{https://doi.org/#1}{\nolinkurl{#1}}}
\newcommand{\arxiv}[1]{arXiv:\href{https://arxiv.org/abs/#1}{\nolinkurl{#1}}}
\newenvironment{funding}{\section*{Funding}}{}
\usepackage{microtype}
\usepackage{xcolor}
\usepackage{placeins}
\newtheorem{theorem}{Theorem}[section]
\newtheorem{proposition}[theorem]{Proposition}
\newtheorem{example}[theorem]{Example}
\newtheorem{lemma}[theorem]{Lemma}

\theoremstyle{definition}

\newtheorem{remark}[theorem]{Remark}

\newcommand{\R}{\mathbb R}

\newcommand{\E}{\mathbb E}
\newcommand{\Pp}{\mathbb P}
\newcommand{\dist}{\operatorname{dist}}
\newcommand{\cone}{\operatorname{cone}}
\newcommand{\conv}{\operatorname{conv}}
\newcommand{\relu}{\operatorname{ReLU}}
\newcommand{\cl}{\operatorname{cl}}
\newcommand{\ip}[2]{\langle #1,#2\rangle}
\newcommand{\norm}[1]{\lVert #1\rVert}

\newcommand{\Unif}{\operatorname{Unif}}
\newcommand{\He}{\operatorname{He}}
\newcommand{\one}{\mathbf 1}
\newcommand{\crosscomplex}{w\!\left(\cl\bigcup_{i,j}(K_{ij}\cap S^{d-1})\right)}

\begin{document}

\title{Two-Layer Random ReLU Networks Can Separate Disjoint Classes}
\author{%
  Yichen Liu\thanks{Email: \texttt{yichen-liu@hhu.edu.cn}.}
  \quad
  Jianbin Yang\thanks{Corresponding author. Email: \texttt{jbyang@hhu.edu.cn}.}\\[0.5em]
  {\normalsize Department of Mathematics, \href{https://ror.org/01wd4xt90}{Hohai University}}\\
  {\normalsize 8 Focheng West Road, Jiangning District}\\
  {\normalsize Nanjing 211100, China}
}
\date{}
\maketitle

\begin{abstract}

Set separation by continuous linear functionals is fundamental in functional analysis and convex optimization and underlies binary classification in machine learning. Support vector machines provide maximum-margin affine separators for linearly separable data, while kernel methods represent nonlinear decision rules through linear separation in a reproducing-kernel Hilbert space. We study when two classes, not necessarily convex or linearly separable in the input space, can be separated with positive margin by finite ReLU networks and by random ReLU features.

At the deterministic level, we prove that positive-margin separation by a finite ReLU network is equivalent to the existence of finite closed convex covers whose opposite-label members have positive pairwise distances, and exactly two hidden layers suffice. 
For random separation, suppose each class is represented as a finite union of nonempty closed convex cells, possibly unbounded. If every opposite-label cell pair has positive distance and its closed difference cone is pointed, then two independent random ReLU layers yield a common positive-margin affine separator with high probability. 

The first layer separates the feature image of each negative cell
from the feature image of the positive class by a hyperplane
through the origin.
It also controls all cross-class distances uniformly. The second layer combines cellwise scores through Chebyshev polynomials and realizes the resulting polynomial using Gaussian Hermite features, followed by a uniform finite-width approximation. With the remaining geometric, bias, and confidence parameters fixed, each sufficient width is quadratic in the largest Gaussian width of a spherical difference-cone section, hence at most linear in its span dimension. A linearly inseparable construction based on translated circular cones
and singletons shows that the squared Gaussian-width dependence
is sharp for both layers.

\end{abstract}



\section{Introduction}

Separating two sets by continuous linear functionals is a basic problem in functional analysis and convex optimization. Under appropriate hypotheses, separation theorems turn geometric information such as disjointness or positive distance into linear certificates for duality and optimality \cite{rockafellar1970}. In machine learning, the same principle underlies binary classification. When input features are linearly separable, a support vector machine selects a maximum-margin affine separator \cite{cortes1995}. When linear separation fails, a kernel method can map the data into a reproducing-kernel Hilbert space (RKHS), where a linear separator gives a nonlinear decision rule in the original space \cite{scholkopf2002}. This approach is powerful, but its geometry depends on the chosen kernel and can entail substantial computational and storage costs. These limitations motivate a complementary question: when can a fixed random network feature map separate two entire classes, including classes that are unbounded or not linearly separable in the input space?

Random neural networks offer one approach to this question. Their hidden weights are sampled once, producing nonlinear features on which a linear readout can be fitted; random-feature methods use this principle for scalable classification and regression \cite{rahimi2007}. ReLU is especially natural because it is positively homogeneous and widely used in neural networks. Recent work has studied whether random ReLU layers can create linear separability without training the hidden weights \cite{dirksen2022,ghosal2022}. We study separation of whole classes rather than only finite samples. The classes themselves need not be convex: finite closed convex covers provide the geometric structure used in our deterministic characterization and random-network analysis.

Our first result gives an exact deterministic criterion. For arbitrary nonempty classes $S^+,S^-\subset\R^d$, a finite ReLU network with an affine output separates them with positive margin if and only if each class is contained in a finite union of nonempty closed convex sets and every pair of opposite-label covering sets has positive distance. Moreover, exactly two hidden layers suffice whenever such covers exist.

We next state the random result. For this theorem, write
\begin{equation}\label{eq:classes}
	S^+=\bigcup_{i=1}^{N_+}E_i^+,
	\qquad
	S^-=\bigcup_{j=1}^{N_-}E_j^-,
\end{equation}
where the cells $E_i^+$ and $E_j^-$ are nonempty, closed, and convex; neither class is required to be convex. Assume that every opposite-label gap
$\delta_{ij}=\dist(E_i^+,E_j^-)$ is positive. For each $i,j$, define the closed difference cone
$K_{ij}=\cl\cone(E_i^+-E_j^-)$. The random separation theorem assumes that each $K_{ij}$ is pointed:
\begin{equation}\label{eq:pointed}
	K_{ij}\cap(-K_{ij})=\{0\}.
\end{equation}
Put $\delta_0=\min_{i,j}\delta_{ij}$, let $R$ and $\kappa_{\min}$ be the scale and cone-depth parameters defined in Section~\ref{sec:main-results}, and set
\[
\omega_*:=\max_{i,j}w(K_{ij}\cap S^{d-1}).
\]
For any $0<\eta<1$, take the second-layer bias radius to be
$\widehat\lambda=\delta_0$. There are explicit factors $\theta_1,\theta_2$, given in Theorems~\ref{thm:first} and~\ref{thm:second}, such that the first-layer bias radius and both widths may be chosen to satisfy
\[
\begin{aligned}
	\lambda&\ge C\delta_0\frac{1+R/\delta_0}{\sqrt{\kappa_{\min}}}
	\sqrt{\log(4N_+)},\\
	n&\ge C\theta_1\left[1+\omega_*^2+\log\frac{48N_+N_-}{\eta}\right],\\
	\widehat n&\ge C\theta_2(1+\omega_*^2).
\end{aligned}
\]
Here $C>0$ is an absolute constant. Under these conditions, with probability at least $1-\eta$ over the two independent random layers, there exist a unit vector $v$, an offset $\tau\in\R$, and a deterministic margin $m>0$ such that
\[
\ip{v}{F(x)}+\tau\ge m\quad(x\in S^+),
\qquad
\ip{v}{F(y)}+\tau\le -m\quad(y\in S^-).
\]
Our construction gives $\tau=0$. In both widths, the dependence on $\omega_*$ is quadratic, hence at most linear in the intrinsic span dimension when the other geometric parameters are fixed;
$\theta_1$ and $\theta_2$ have at most
	singly exponential dependence on the remaining parameters. 

Our contributions are threefold.
\begin{enumerate}[label=(\arabic*),leftmargin=2.2em]
	\item \textbf{A deterministic characterization of set separation.}
	Positive-margin separation by a finite ReLU network is equivalent to the existence of finite closed convex covers with positive distances between all opposite-label cover sets; exactly two hidden layers suffice (Theorem~\ref{thm:deterministic}). We also give an explicit unbounded example for which no finite-width one-hidden-layer ReLU network achieves strict separation (Example~\ref{ex:shallow}).
	
	\item \textbf{Random two-layer separation beyond bounded data.}
	For an individual pair of convex covering cells at positive distance, pointedness of the difference cone is equivalent to positive probability of strict Gaussian affine separation with a suitable finite uniform bias range; cone depth quantifies the angular margin (Proposition~\ref{thm:cone}). Under positive opposite-cell gaps and pairwise pointedness, the first random layer separates the image of each negative cell from the image of the entire positive class and controls all cross-class distances uniformly.   A second independent layer combines these cellwise separators into one positive-margin separator for the full classes. The proof uses normalized endpoints, Chebyshev polynomials, and Gaussian Hermite features. A nonpointed strip example shows that positive gaps alone do not ensure random separation at any fixed finite depth (Example~\ref{ex:nonpointed-depth}).
	
\item \textbf{Intrinsic width bounds and sharpness.}
Each sufficient width is quadratic in the largest Gaussian width
of a spherical difference-cone section, and this order is necessary
for both layers on a linearly inseparable family of
	translated circular cones and singletons
(Example~\ref{ex:numerator-lower}).
Thus the width dependence is governed by intrinsic cone geometry
rather than ambient dimension, while both prefactors
	have at most single-exponential dependence on the remaining parameters.
Numerical experiments illustrate the width trends for
non-linearly-separable classes with convex components.
\end{enumerate}

Random-feature methods have been studied for kernel approximation and learning \cite{rahimi2007}, and recent work gives random-ReLU separation guarantees for bounded data and quadratic-feature guarantees for unions of subspaces \cite{dirksen2022,ghosal2022,xu2026}. A detailed comparison of these results with our theorems, along with related geometric-embedding results, appears in Subsection~\ref{subsec:comparison}.

The remainder of the paper is organized as follows. Section~\ref{sec:main-results} states the deterministic characterization, the geometric criterion for random affine separation, and the two-layer random separation theorem. Section~\ref{sec:extensions} gives the pointedness obstruction and matching width lower bounds, and compares these results with related work. Section~\ref{sec:experiments} reports the numerical experiments. The main text closes with concluding remarks; the appendices contain the proofs.

\section{Two-Layer ReLU Separation of Classes}\label{sec:main-results}

This section develops separation results for two labeled sets, which need not themselves be convex. We first give the deterministic finite-network characterization, discuss the limitations of one hidden layer, and introduce the problem considered in this paper. We then state the geometric criterion for random affine separation and the high-probability results for two independent random ReLU layers.

Let $\norm x=(\ip xx)^{1/2}$, and let $P_A$ denote the Euclidean projection
onto a closed convex set $A$. We write $\cone A$, $\cl A$, and $\conv A$ for the conic hull, closure, and convex hull of a set $A$, respectively. Define the closed unit ball and the unit sphere in $\R^d$ by
\[
B_2^d=\{u\in\R^d:\norm u\le1\},
\qquad
S^{d-1}=\{u\in\R^d:\norm u=1\}.
\] For each positive integer $N$, write
$[N]=\{1,\ldots,N\}$. For a nonempty bounded set $A$, define its Gaussian width by
\[
w(A)=\E_g\sup_{z\in A}\ip gz,
\qquad
g\sim N(0,I),
\]
where $g$ is a standard Gaussian vector in the ambient space of $A$.

 For a real random variable
$X$, a Euclidean random vector $Y$, and $1\le p<\infty$, define
\[
\norm{X}_{L^p}=(\E_X|X|^p)^{1/p},
\qquad
\norm{Y}_{L^p}=(\E_Y\norm Y^p)^{1/p}.
\]
The inner norm $\norm Y$ is Euclidean, and the $L^p$ norm of a function is
taken with respect to the random argument specified by the context. For
matrices, $\norm{\cdot}_{\mathrm F}$ denotes the Frobenius norm.

\subsection{Network model and separation criteria}
The criterion below applies to arbitrary nonempty sets. It identifies exactly when a finite ReLU network can separate them with positive margin and shows that two hidden layers always suffice.
\begin{theorem}[Finite convex-cover characterization]\label{thm:deterministic}
	For nonempty sets $S^+,S^-\subset\R^d$, the following are equivalent.
	\begin{enumerate}[label=(\roman*)]
		\item A finite ReLU network followed by an affine output separates $S^+$ and
		$S^-$ with positive margin.
		\item There exist finite families of nonempty closed convex sets
		$\{C_i^+\}_{i=1}^{M_+}$ and $\{C_j^-\}_{j=1}^{M_-}$ such that
		\[
		S^+\subset\bigcup_i C_i^+,\qquad
		S^-\subset\bigcup_j C_j^-,\qquad
		\dist(C_i^+,C_j^-)>0\quad\text{for all }i,j.
		\]
		\item A ReLU network with exactly two hidden layers followed by an affine
		output separates $S^+$ and $S^-$ with positive margin.
	\end{enumerate}
\end{theorem}
The proof is in Appendix~\ref{app:necessary}.

For bounded classes, the cover condition in Theorem~\ref{thm:deterministic} is equivalent to $\dist(S^+,S^-)>0$, and a finite-width one-hidden-layer ReLU network also suffices \cite{ghosal2022}.
For unbounded classes, one hidden layer may fail, as the example below shows.

\paragraph{Why one hidden layer may fail.}

Let $s=(s_1,s_2,s_3)\in\{-1,1\}^3$ and define
\begin{equation}\label{eq:orthant-cells}
	E_s=\{x\in\R^3:s_rx_r\ge1,\ r=1,2,3\},
	\qquad \chi(s)=s_1s_2s_3.
\end{equation}
The two classes in Example~\ref{ex:shallow} are
\begin{equation}\label{eq:orthant-classes}
	S^+=\bigcup_{\chi(s)=1}E_s,
	\qquad
	S^-=\bigcup_{\chi(s)=-1}E_s.
\end{equation}
The two classes have distance $2$.

\begin{example}[Shallow obstruction]\label{ex:shallow}
	For the classes $S^+,S^-$ in \eqref{eq:orthant-classes}, with cells
	from \eqref{eq:orthant-cells}, no finite-width one-hidden-layer ReLU function
	\[
	h(x)=c+\sum_{k=1}^m a_k\relu(\ip{v_k}{x}+b_k)
	\]
	has opposite signs with a uniform positive margin on $S^+$ and $S^-$.  The same
	is true for strict zero-threshold separation.  The example embeds into every
	$\R^d$, $d\ge3$.
\end{example}

The proof is in Appendix~\ref{app:necessary}.

\begin{remark}[Necessity of two hidden layers]
The example above  shows that one hidden layer can fail even when the deterministic finite-cover criterion holds. By Theorem~\ref{thm:deterministic}, two hidden layers suffice for this example. Thus, two hidden layers are necessary for some class pairs. 
\end{remark}

In the following, 
we consider the composition $F=\widehat\Phi\circ\Phi$ of two
independent random ReLU layers,

\begin{equation}\label{eq:network}
	\Phi(x)=\sqrt{\frac{2}{n}}\,\relu(Wx+b),
	\qquad
	\widehat\Phi(z)=\sqrt{\frac{2}{\widehat n}}\,
	\relu(\widehat Wz+\widehat b),
\end{equation}
where $W\in\R^{n\times d}$ and
$\widehat W\in\R^{\widehat n\times n}$ have standard Gaussian entries.
The bias vectors $b\in\R^n$ and $\widehat b\in\R^{\widehat n}$
have entries uniform on $[-\lambda,\lambda]$ and
$[-\widehat\lambda,\widehat\lambda]$, respectively.
All matrix and vector entries are mutually independent.

This sampling scheme follows \cite[Definition~1]{dirksen2022}.
The factor $\sqrt{2}$ compensates for the second-moment reduction
of ReLU under symmetric inputs, as in random initialization
\cite{he2015}. Here each layer is normalized by the square root of its output width.

\paragraph{Two-layer random ReLU separation problem.}
Given the classes in \eqref{eq:classes} and a failure probability $\eta\in(0,1)$, assume that $\delta_0>0$ and every difference cone in \eqref{eq:pointed} is pointed. We seek conditions on the widths $n,\widehat n$ and bias scales $\lambda,\widehat\lambda$ that guarantee the following: there is a margin $m>0$, fixed before sampling, such that with probability at least $1-\eta$ the sampled network admits a unit vector $v\in\R^{\widehat n}$ and an offset $\tau\in\R$ satisfying
\[
\ip{v}{F(x)}+\tau\ge m\quad(x\in S^+),
\qquad
\ip{v}{F(y)}+\tau\le -m\quad(y\in S^-).
\]

The next result concerns the random directions used in our class-separation theorem. For an individual pair of convex covering cells, positive distance guarantees deterministic affine separation, but a Gaussian draw can miss all strictly separating directions with probability one when the cells are unbounded. The parameter $\kappa(K)$ measures the robustness of a separating direction.
A larger value gives greater tolerance to changes in direction.
It also enters the bounds on separation probabilities and sufficient
widths for random ReLU networks.
\begin{proposition}[Cone geometry of affine separation]\label{thm:cone}
	Let $E^+ , \, E^-\subset\R^d$ be nonempty closed convex sets with positive distance.
	Put $K=\cl\cone(E^+-E^-)$. Define
	\[
	C_K=\conv(K\cap S^{d-1}),\qquad
	\kappa(K)=\dist(0,C_K).
	\]
	For $a\ne0$, the affine hyperplane $\ip ax+c=0$ linearly separates
	$E^-$ and $E^+$ with  margin $t>0$ if
	\[
	\ip ax+c\le-t\quad(x\in E^-),\qquad
	\ip ax+c\ge t\quad(x\in E^+).
	\]
	When $\norm a=1$, the parameter $t$ is a geometric margin.
	The following assertions hold.
	\begin{enumerate}[label=(\roman*)]
		\item $K$ is pointed if and only if there exist $\lambda,t>0$ such that, with positive probability, the random affine hyperplane
		$\ip gx+b=0$, where $g\sim N(0,I_d)$ and
		$b\sim\mathrm{Unif}[-\lambda,\lambda]$ are independent, separates
		$E^-$ and $E^+$ with margin $t$.
		
		\item If $K$ is pointed,  set $u_K=P_{C_K}(0)/\norm{P_{C_K}(0)}$.  Then $\kappa(K)>0$, and $u_K$ is the unique solution of
		\[
		\max_{u\in S^{d-1}}\inf_{z\in K\cap S^{d-1}}\ip uz
		=\kappa(K).
		\]
		In particular,
		\begin{equation}\label{eq:depthineq}
			\ip{u_K}{z}\ge\kappa(K)\norm z
			\qquad(z\in K).
		\end{equation}
	\end{enumerate}
\end{proposition}
The proof is in Appendix~\ref{app:necessary}.

\begin{figure}[H]
	\centering
	\includegraphics[width=.72\linewidth]{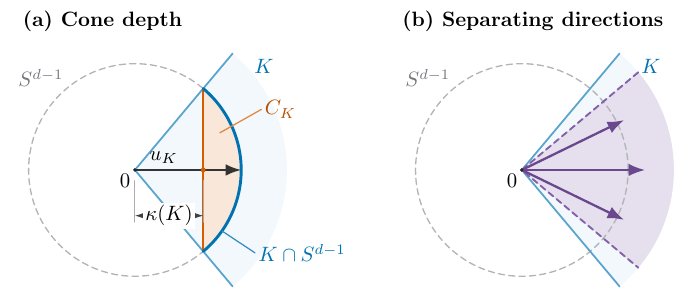}
	\caption{Cone depth (a). The purple sector in (b) shows separating
		directions; its dashed boundary rays are excluded.}
	\label{fig:kappa-geometry}
\end{figure}

\subsection{Random two-layer separation of the classes}

We now state the random separation theorem. To quantify the geometry of the opposite-label cells, define
\[
x_i^\circ=P_{E_i^+}(0),\qquad
y_j^\circ=P_{E_j^-}(0),\qquad
R=\max_{i,j}\{\norm{x_i^\circ},\norm{y_j^\circ}\}.
\]
The parameter $R$ records the scale of the nearest points of these cells
to the origin and remains finite.
Under \eqref{eq:pointed}, write
$\kappa_{ij}=\kappa(K_{ij})$, $u_{ij}=u_{K_{ij}}$, and
$\kappa_{\min}=\min_{i,j}\kappa_{ij}>0$ by Proposition~\ref{thm:cone}(ii).

\begin{figure}[!htbp]
	\centering
	\includegraphics[width=\linewidth]{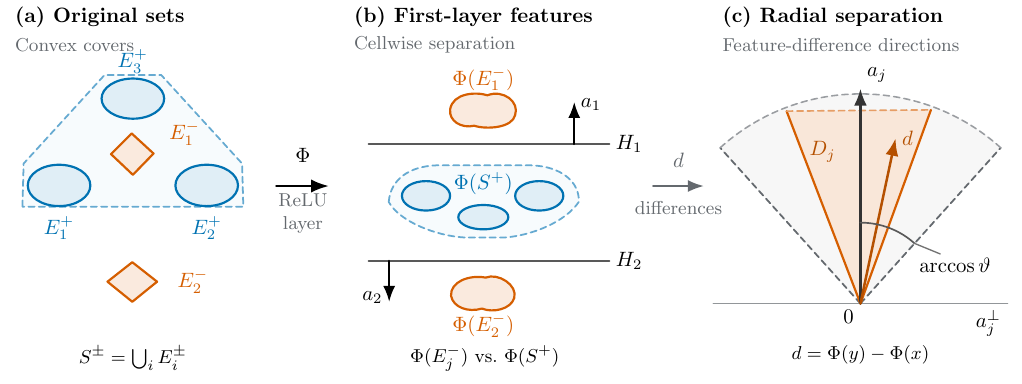}
	\caption{First-layer ReLU transformation and radial separation
		(Theorem~\ref{thm:first}).
		(a) The two classes are represented by positive and negative convex cells.
		(b) The map $\Phi$ produces feature cells for which each $\Phi(E_j^-)$ is
		separated from the entire $\Phi(S^+)$ by $H_j$, with normal $a_j$ pointing
		towards the negative cell.  Dashed blue outlines enclose positive convex hulls.
		(c) For a fixed $j$, the difference cone
		$D_j=\cl\cone(\Phi(E_j^-)-\Phi(S^+))$ lies in the gray angular envelope:
		every nonzero $d\in D_j$ satisfies
		$\angle(a_j,d)\le\arccos\vartheta<\pi/2$.
		The origin is that of the difference space.}
	\label{fig:first-layer-radial}
\end{figure}

The first random ReLU layer separates each $\Phi(E_j^-)$ from $\Phi(S^+)$, as illustrated in Figure~\ref{fig:first-layer-radial}. 
The next theorem gives sufficient width and bias conditions and a uniform margin that the second layer uses to separate the two classes as a whole.

\begin{theorem}[First-layer separation]\label{thm:first}
	Given the classes in \eqref{eq:classes}, assume that \eqref{eq:pointed} holds and $\delta_0>0$.
	Let $0<\eta<1$. There exists a positive constant $C$, independent of $S^+$, $S^-$ and $\eta$, such that if
	\begin{equation}\label{eq:first-bias}
		\lambda\ge C\delta_0
		\frac{1+R/\delta_0}{\sqrt{\kappa_{\min}}}\sqrt{\log(4N_+)}
	\end{equation}
	and
	\begin{equation}\label{eq:first-width}
		n\ge C\theta_1\left[1+\max_{i,j}w(K_{ij}\cap S^{d-1})^2
		+\log\frac{48N_+N_-}{\eta}\right]  ,
	\end{equation}
	then, with probability at least $1-\eta$, there exist a common constant $\vartheta>0$ and unit vectors $a_j\in\R^n$,  $j\in[N_-]$, such that, simultaneously for every $j\in[N_-]$, $x\in S^+$, and $y\in E_j^-$,
	\begin{equation}\label{eq:first-angle}
		\ip{a_j}{\Phi(y)-\Phi(x)}
		\ge\vartheta\norm{\Phi(y)-\Phi(x)},
	\end{equation}
	and
	\begin{equation}\label{eq:first-bilip}
		\vartheta\norm{x-y}\le\norm{\Phi(x)-\Phi(y)}
		\le\tfrac32\norm{x-y}.
	\end{equation}
	In particular, every $\cl\cone(\Phi(E_j^-)-\Phi(S^+))$ is pointed. Here
	\[
	\theta_1=\frac\lambda{\delta_0}
	\exp\!\left\{\frac{C(1+R/\delta_0)^2\log(4N_+)}{\kappa_{\min}}\right\}.
	\]
	The constant $\vartheta$ can be determined from the parameters chosen before sampling.
\end{theorem}

The proof is in Appendix~\ref{app:first}. 

The second layer $\widehat\Phi$ takes the first-layer features
$\Phi(x)$, $x\in S^+\cup S^-$, as its inputs. The following proposition
shows that $\norm{\Phi(x)}$ is comparable to
$\bigl(\norm{x}^2+\lambda^2\bigr)^{1/2}$. These norm estimates are used
to derive the parameter bounds for the second layer.

\begin{proposition}[Bounds on $\norm{\Phi(x)}$]\label{prop:first-norm}
	Under the assumptions of Theorem~\ref{thm:first}, with probability at
	least $1-\eta$, the conclusions of that theorem hold together with the following bounds, uniformly for all
	$x\in S^+\cup S^-$:
	\begin{equation}\label{eq:full-norm}
		\frac18\sqrt{\norm{x}^2+\lambda^2}
		\le \norm{\Phi(x)}
		\le 3\sqrt{\norm{x}^2+\lambda^2}.
	\end{equation}
\end{proposition}

The proof is in Appendix~\ref{app:first}.

\begin{figure}[!htbp]
	\centering
	\includegraphics[width=.35\linewidth]{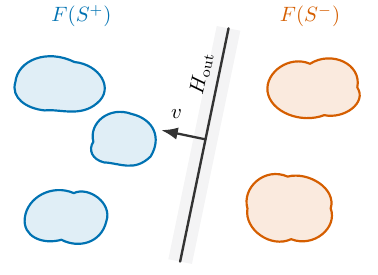}
	\caption{Linear separation after two random ReLU layers.
		The feature sets $F(S^+)$ and $F(S^-)$ are separated with positive margin
		by $H_{\mathrm{out}}=\{z:\ip{v}{z}=0\}$.
		The normal $v$ points towards $F(S^+)$.}
	\label{fig:second-layer-separation}
\end{figure}

The following theorem establishes our main random separation result. The first layer separates each feature image $\Phi(E_j^-)$ from $\Phi(S^+)$. The second layer then makes the full feature sets $F(S^+)$ and $F(S^-)$ linearly separable with a positive margin, as illustrated in Figure~\ref{fig:second-layer-separation}.

\begin{theorem}[Second-layer separation]\label{thm:second}
	Assume the setting of Theorem~\ref{thm:first}, and suppose that the
	first-layer conditions \eqref{eq:first-bias}--\eqref{eq:first-width}
	hold. Set $\widehat\lambda=\delta_0$. There exists a universal constant
	$C>0$, independent of the sets and $\eta$, such that the following
	holds. If
	\begin{equation}\label{eq:second-width}
		\widehat n\ge C\theta_2
		\left[1+\max_{i,j}w(K_{ij}\cap S^{d-1})^2\right],
	\end{equation}
	then with probability at least $1-\eta$, there exist a margin $t>0$
	and a unit vector $v\in\R^{\widehat n}$ such that
	\begin{equation}\label{eq:final-sign}
		\ip v{F(x)}\ge t\quad(x\in S^+),\qquad
		\ip v{F(y)}\le-t\quad(y\in S^-).
	\end{equation}
	Here
	\[
	\theta_2=\exp\!\left\{
	\frac{CN_-(1+R/\delta_0)^2}{\kappa_{\min}}
	\log\!\left(\frac{4N_+N_-\lambda}{\delta_0}\log\frac2\eta\right)
	\right\}.
	\]
 	On the same event, the conclusions of Theorem~\ref{thm:first} and
Proposition~\ref{prop:first-norm} also hold. The margin $t$ depends
only on parameters fixed before sampling, whereas $v$ may depend on
the sampled network.
\end{theorem}
The proof of Theorem~\ref{thm:second} is given in
Appendix~\ref{app:second}.

The theorem applies to disjoint classes,
which may be unbounded, and provides sufficient widths for their
separation with probability at least $1-\eta$. Fix the remaining
geometric parameters and $\eta$, and define the intrinsic dimension
\[
r:=\max_{i,j}\dim\operatorname{span}(K_{ij}).
\]
For each $i,j$, if $g\sim N(0,I_d)$, then
\[
w(K_{ij}\cap S^{d-1})
\le \E_g\bigl\|P_{\operatorname{span}(K_{ij})}g\bigr\|
\le \sqrt{\dim\operatorname{span}(K_{ij})}
\le \sqrt r,
\]
where the second inequality follows from the Cauchy--Schwarz
inequality. Hence
\[
\max_{i,j}w(K_{ij}\cap S^{d-1})^2\le r,
\]
so both sufficient width bounds grow at most linearly with $r$ when
the remaining parameters are fixed.

\section{Geometric Obstructions and Sharp Width Bounds}\label{sec:extensions}
This section examines the necessity of pointedness and the dependence
of network width on Gaussian complexity.
We first construct classes at positive distance with nonpointed
difference cones. Every fixed finite-depth, finite-width random ReLU
network fails to separate these classes almost surely.
We then use translated circular cones and singletons to derive necessary widths
for both layers. These lower bounds match the squared Gaussian-width
dependence in our sufficient bounds.
The section concludes with a comparison of our results with related work
 
\subsection{Necessity of the pointedness assumption}
\label{subsec:pointedness-necessity}
Positive opposite-cell gaps alone do not guarantee random separation. Example~\ref{ex:nonpointed-depth} exhibits classes with nonpointed difference cones for which every fixed finite-depth, finite-width random ReLU network fails to separate the classes almost surely. Thus, a positive gap condition alone cannot yield a uniform random separation guarantee.

  Consider the planar convex cover
\begin{equation}\label{eq:nonpointed-strips}
\begin{gathered}
E_1^+=[0,\infty)\times[1,2],\qquad
E_2^+=[0,\infty)\times[-2,-1],\\
E_1^-=[0,\infty)\times[-\tfrac12,\tfrac12],\qquad
S^+=E_1^+\cup E_2^+,\quad S^-=E_1^-.
\end{gathered}
\end{equation}
The three cells are closed and convex with nonempty interiors.  Both
opposite-cell gaps equal $1/2$, whereas
\[
K_{11}=\R\times[0,\infty),\qquad
K_{21}=\R\times(-\infty,0].
\]
Both difference cones contain the line $\R e_1$, so neither is pointed and
$\kappa(K_{11})=\kappa(K_{21})=0$.

\begin{example}[Random separation fails at every finite depth]
\label{ex:nonpointed-depth}
For the classes in \eqref{eq:nonpointed-strips}, fix any finite depth $q\ge1$,
widths $n_1,\ldots,n_q\ge1$, and finite bias ranges
$\lambda_1,\ldots,\lambda_q>0$.  Let $F_0(x)=x$ and
\[
F_\ell(x)=\sqrt{\frac{2}{n_\ell}}\,
\relu\!\left(W_\ell F_{\ell-1}(x)+b_\ell\right),
\qquad 1\le\ell\le q,
\]
where all weight entries are independent $N(0,1)$ variables and all bias
entries are independent uniform variables on their respective
$[-\lambda_\ell,\lambda_\ell]$, independently of the weights.
Then, almost surely,
\begin{equation}\label{eq:nonpointed-hull}
F_q(S^-)\cap\conv\bigl(F_q(S^+)\bigr)\ne\varnothing.
\end{equation}
Consequently, no affine readout separates the two classes for almost every network realization, even if the readout is chosen after observing the network. In particular, positive-margin separation fails almost surely.
\end{example}
The proof is in Appendix~\ref{app:necessary}.

\begin{figure}[htbp]
\centering
\includegraphics[width=0.8\linewidth]{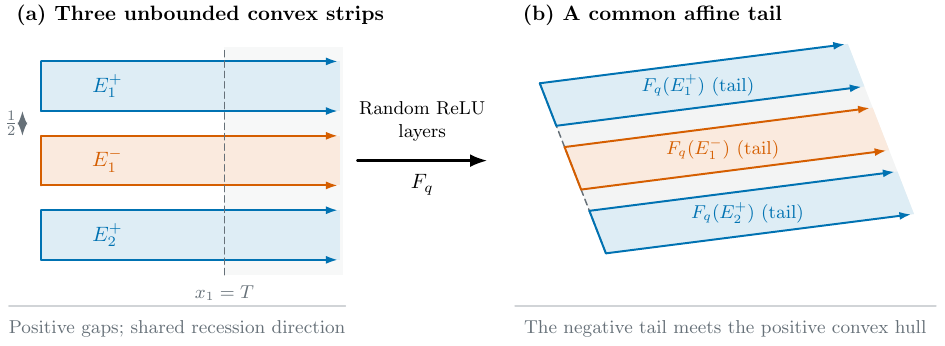}
\caption{A planar obstruction to random separation at every finite depth
(Example~\ref{ex:nonpointed-depth}).
(a) Two positive strips surround a negative strip, with opposite-cell gaps
$1/2$ and a common recession direction.
(b) Almost surely, the network is affine on
$\{(t,s):t\ge T,\ |s|\le2\}$ for some finite $T$; a nondegenerate affine
image is shown.  In particular,
$F_q(t,0)=[F_q(t,1)+F_q(t,-1)]/2$, so the negative image meets the convex
hull of the positive image.}
\label{fig:pointedness-necessity}
\end{figure}

The deterministic two-neuron score
$h(x)=\relu(x_2)+\relu(-x_2)-3/4$ satisfies $h\ge1/4$ on $S^+$ and
$h\le-1/4$ on $S^-$.  Random hidden layers encounter a different obstruction:
on a sufficiently distant common tail, their composition is affine and
preserves a cross-class midpoint relation.  Figure~\ref{fig:pointedness-necessity}
illustrates this geometry; 

\subsection{Necessity of the squared Gaussian width}\label{sec:necessary}
We now give a family of sets that attains a lower bound of this squared order with fixed geometric parameters.
Let $r\ge2$, $0<\alpha<\pi/2$, and define
\begin{equation}\label{eq:optimal-circular-cone}
\begin{gathered}
K_{\alpha,r}=\{z\in\R^r:z_1\ge\cos\alpha\,\norm z\},\\
E_1^+=\tfrac32 e_1+K_{\alpha,r},\qquad
E_1^-=-\tfrac32e_1-K_{\alpha,r},\\
E_2^+=\{-\tfrac12e_1\},\qquad E_2^-=\{\tfrac12e_1\},
\qquad S^\pm=E_1^\pm\cup E_2^\pm.
\end{gathered}
\end{equation}
Each class consists of a translated circular cone and a singleton.
Both convex hulls contain $\{te_1:|t|\le1/2\}$, so the classes are not
linearly separable in the input space. They satisfy
\begin{equation}\label{eq:lower-family-geometry}
\begin{gathered}
N_+=N_-=2,\qquad \delta_0=1,\qquad R=\tfrac32,
\qquad \kappa_{\min}=\cos\alpha,\\
K_{11}=K_{12}=K_{21}=K_{\alpha,r},\qquad
K_{22}=\{-te_1:t\ge0\}.
\end{gathered}
\end{equation}
The spherical section of $K_{22}$ is a singleton and has Gaussian width zero. Hence
\[
\max_{i,j}w(K_{ij}\cap S^{r-1})^2
=w(K_{\alpha,r}\cap S^{r-1})^2.
\]
For $G_{r-1}\sim N(0,I_{r-1})$, the Gaussian width of the spherical section satisfies
\begin{equation}\label{eq:circular-width-bounds}
\sin\alpha\,\E_{G_{r-1}}\norm{G_{r-1}}
\le w(K_{\alpha,r}\cap S^{r-1})
\le\tfrac32\sin\alpha\,\E_{G_{r-1}}\norm{G_{r-1}}.
\end{equation}
Fix $\alpha$ and increase $r$. The other geometric parameters remain unchanged, while the squared Gaussian width is of order $r$.

\begin{example}[Necessary width of each layer]\label{ex:numerator-lower}
For~\eqref{eq:optimal-circular-cone}, consider a two-layer ReLU network with independent Gaussian weights.
The biases may take any finite values. Let $0<\eta<1$.
If $F(S^+)$ and $F(S^-)$ are strictly linearly separable with probability at least $1-\eta$,
then each layer must satisfy
\begin{equation}\label{eq:both-width-necessary}
n,\widehat n\ge(1-\eta)r\sin^2\alpha
\ge\tfrac49(1-\eta)w(K_{\alpha,r}\cap S^{r-1})^2.
\end{equation}
If only the first-layer radial separation in Theorem~\ref{thm:first} is required with probability at least $1-\eta$,
then $n\ge(1-\eta)r\sin^2\alpha$ is still necessary.
\end{example}

The proof is in Appendix~\ref{app:necessary}. The obstruction is a pair of points from different classes with identical features.
Thus adding any constant to $\ip v{F(x)}$ cannot separate the classes.
Take $\alpha=\pi/4$ and $\eta=1/4$. Fix $\lambda$ as a sufficiently large absolute constant according to~\eqref{eq:first-bias},
and take $\widehat\lambda=1$.
Then
\[
\frac{r-1}{\pi}\le w(K_{\pi/4,r}\cap S^{r-1})^2
\le\frac98(r-1).
\]
Theorems~\ref{thm:first} and~\ref{thm:second} give a sufficient width of
$C[1+w(K_{\pi/4,r}\cap S^{r-1})^2]$ for each layer.
Equation~\eqref{eq:both-width-necessary} and $1+\frac98(r-1)\le\frac98r$ give
\[
n,\widehat n\ge\frac{3r}{8}
\ge\frac13[1+w(K_{\pi/4,r}\cap S^{r-1})^2].
\]
This shows that, for independent Gaussian random hidden layers, the squared Gaussian-width order cannot be reduced in uniform width bounds over this family of sets. The lower bound constrains each layer separately; widening one layer cannot remove the restriction on the other.

\subsection{Comparison of results and related work}\label{subsec:comparison}
This subsection compares our results with related work on random
neural networks and compressive classification.

\paragraph{Linear separability of shallow features.}
Studies of trained networks show that, as depth increases,
features can become more linearly separable and features within
each class can cluster more tightly
\cite{alain2016,he2023,masarczyk2023}.
For random networks, earlier results establish separation
of bounded classes using one or two ReLU layers
\cite{dirksen2022,ghosal2022}.  Quadratic random features have also been studied for union-of-subspaces
models \cite{xu2026}.
	Related work establishes generalization rates for deep ReLU classifiers
	with logistic loss and optimization and generalization guarantees for
	minimal-width shallow networks
	\cite{zhang2024logloss,lei2026minimalwidth}.
	Approximation results give optimal width-depth rates for H\"older functions
	and nearly optimal bounds for smooth functions
	\cite{shen2022widthdepth,lu2021smooth}.
	We study entire-class separation by fixed random ReLU layers,
	including on unbounded sets, with widths controlled by difference-cone geometry.
Our result allows finite unions of closed convex sets,
provided that opposite-set gaps are positive and the difference cones
are pointed.
The first layer separates each $\Phi(E_j^-)$ from
$\Phi(S^+)$. The second layer then makes $F(S^+)$ and $F(S^-)$
linearly separable with a positive margin.
With the remaining parameters fixed, the sufficient width of each
layer is quadratic in the Gaussian widths of the spherical
difference-cone sections and at most linear in their intrinsic span
dimension.
Table~\ref{tab:separation-comparison} gives the precise set assumptions,
network models, and separation guarantees.

\begin{table}[!htbp]
	\centering
	\caption{High-probability separation by random features.
		Widths are sufficient bounds with remaining geometric and covering
		parameters and failure probability fixed; bias scales satisfy the cited
		bounds. In the first two rows, intrinsic dimension is the span dimension
		of the input classes. Weights are Gaussian; ReLU layers have uniform
		biases and are independent.}
	\label{tab:separation-comparison}
	\setlength{\tabcolsep}{3pt}
	\renewcommand{\arraystretch}{1.04}
	\begin{tabular}{@{}>{\raggedright\arraybackslash}p{\dimexpr.15\linewidth-1.5\tabcolsep\relax}
			>{\raggedright\arraybackslash}p{\dimexpr.27\linewidth-1.5\tabcolsep\relax}
			>{\raggedright\arraybackslash}p{\dimexpr.15\linewidth-1.5\tabcolsep\relax}
			>{\raggedright\arraybackslash}p{\dimexpr.43\linewidth-1.5\tabcolsep\relax}@{}}
		\toprule
		Work & Sets & Network & Separation and sufficient width \\
		\midrule
		Dirksen et~al.\newline\cite{dirksen2022}
		& Two bounded sets at positive distance
		& Two ReLU layers.
		& Positive margin.
		First width: quadratic in Gaussian widths, at most linear in intrinsic dimension.
		Second: exponential in squared Gaussian widths, at most exponential in dimension. \\
		\addlinespace[3pt]
		Ghosal et~al.\newline\cite{ghosal2022}
		& Two bounded sets at positive distance
		& One ReLU layer.
		& Positive margin. Width exponential in squared Gaussian width,
		at most exponential in intrinsic dimension; additional mutual-complexity
		dependence (Theorem~1.4). \\
		\addlinespace[3pt]
		Xu et~al.\newline(UoS)\newline\cite{xu2026}
		& Two $r$-dimensional linear subspaces intersecting only at $0$.
		& One quadratic layer; zero bias.
		& Strict separation off $0$; no uniform positive margin.
		Width $O(r^3\log(r/\eta))$ for fixed minimum principal angle. \\
		\addlinespace[3pt]
		This work\newline
		Theorems\newline
		\ref{thm:first}, \ref{thm:second}
		& Finite unions of closed convex sets, possibly unbounded;
		positive opposite-set gaps and pointed difference cones.
		& Two ReLU layers.
		& Positive margin. Each width: quadratic in cone Gaussian widths,
		at most linear in intrinsic dimension; sharp complexity order
		(Example~\ref{ex:numerator-lower}). \\
		\bottomrule
	\end{tabular}
\end{table}
\FloatBarrier
\paragraph{Compressive classification and cone geometry.}

Compressive classification uses compressed measurements directly,
without first reconstructing the signal \cite{davenport2010}.
Bandeira et~al.\ study the rare eclipse problem \cite{bandeira2017}.
It asks how Gaussian projections can reduce dimension while keeping
two disjoint closed convex sets disjoint with high probability.
Using Gordon's escape through a mesh theorem \cite{gordon1988},
they bound sufficient measurement counts in terms of the squared
Gaussian width of the spherical difference-cone section.
In convex recovery, exact recovery is equivalent to the measurement
kernel intersecting the tangent cone at the target signal only
at the origin \cite{chandrasekaran2012}.
Sufficient Gaussian measurement counts are likewise controlled
by the squared Gaussian width of this cone's spherical section.

Our width lower bounds also connect random kernels
	with cone geometry. Here the kernel is the nullspace of a random
	linear map. In Example~\ref{ex:numerator-lower}, if this nullspace
	intersects the interior of the specified cone, we can find two points
	from opposite classes that have identical network features.
	No subsequent layer or classifier can distinguish these points.
	Estimating the probability of this event gives the width lower bounds.
\paragraph{Random-feature methods.}
Random-feature methods and extreme learning machines fix the hidden
parameters after random sampling and learn only the output coefficients
\cite{rahimi2007,huang2006}.
Recent empirical studies show that these methods can achieve test
performance comparable to networks trained end to end on selected
classification and regression benchmarks
\cite{markowska2021,hu2023}.

Random features also admit theoretical guarantees in kernel learning.
For example, under suitable assumptions, ridge regression with random
features can match the learning rate of full kernel ridge regression
\cite{rudi2017}.
For shift-invariant kernels, random features allow linear SVMs to
approximate kernel SVMs without forming the full kernel matrix
\cite{rahimi2007}.
Function approximation theory provides another justification for
fixing the hidden parameters. Under suitable activation and sampling
assumptions, a linear combination of random features can approximate
a target function to a prescribed accuracy with high probability
\cite{needell2024,hsu2021,salanevich2025}.
The sufficient width depends on the target function class, the
required accuracy, and the input dimension. In some bounds, the
dependence on dimension is exponential or superexponential.

Our results give a geometric guarantee for this approach.
With high probability, the two fixed random hidden layers admit an
affine output function that separates the entire classes with a positive
margin. Only the output coefficients need to be chosen after sampling;
the hidden weights and biases remain fixed.

\section[Numerical Experiments on Gaussian-Width Dependence]{Numerical Experiments on Gaussian-Width Dependence}
\label{sec:experiments}

We examine how the Gaussian width in Theorems~\ref{thm:first} and~\ref{thm:second} relates to the widths needed for certified separation. The experiments use four convex cells arranged in an XOR pattern. The first varies Gaussian width while fixing the intrinsic span dimension; the second varies the intrinsic span dimension while keeping Gaussian width nearly fixed.

\subsection{Geometry and experimental setting}

Work in $\R^d$ with $d=1024$. Set $a=1/10$, take $2\le r\le d-1$, and let $0<s\le1$.
\begin{equation*}
	Q_{r,s}=\left\{a\left(u_1e_3+
	s\sum_{\ell=2}^{r-1}u_\ell e_{\ell+2}\right):
	u\in B_2^{r-1}\right\}.
\end{equation*}
Here $B_2^{r-1}$ is the unit Euclidean ball in $\R^{r-1}$. The set $Q_{r,s}$ is an ellipsoidal disk with semiaxis $a$ along $e_3$ and semiaxes $as$ along $e_4,\ldots,e_{r+1}$.
Set
\begin{equation*}
	\begin{array}{ll}
		E_1^+= e_1+\tfrac12e_2+Q_{r,s},&
		E_2^+=-e_1-\tfrac12e_2+Q_{r,s},\\[2pt]
		E_1^-= e_1-\tfrac12e_2+Q_{r,s},&
		E_2^-=-e_1+\tfrac12e_2+Q_{r,s},
	\end{array}
	\qquad S^\pm=E_1^\pm\cup E_2^\pm.
\end{equation*}
Each component is a compact convex ellipsoidal disk of affine dimension $r-1$. For every $q\in Q_{r,s}$, each class contains two points whose midpoint is $q$. Hence $Q_{r,s}\subset\conv(S^+)\cap\conv(S^-)$, so no affine hyperplane strictly separates the classes. Both experiments keep
\begin{equation}\label{eq:xor-fixed-geometry}
	N_+=N_-=2,\qquad \delta_0=1,\qquad R=\frac{\sqrt5}{2},\qquad
	\kappa_{\min}=\frac1{\sqrt{1+4a^2}}.
\end{equation}
The four cell centers are $c=\pm e_1\pm\frac12e_2$, and each is orthogonal to $\operatorname{span}(Q_{r,s})$. Thus the projection of the origin onto each cell is its center, giving $R=\sqrt5/2$. For every opposite-label pair, the center difference is orthogonal to $\operatorname{span}(Q_{r,s})$ and has length $1$ or $2$. Since $0\in Q_{r,s}$, the cell distance equals the center-difference length, and therefore $\delta_0=1$. Each difference cone spans the $r$-dimensional space generated by its center difference and $\operatorname{span}(Q_{r,s})$. The pairs $(E_1^+,E_1^-)$ and $(E_2^+,E_2^-)$ attain $\omega_*$. By rotational invariance of Gaussian width, it suffices to compute this width on the spherical section of $\cl\cone(e_1+2Q_{r,s})$.

We use the network in~\eqref{eq:network} with first-layer bias radius $\lambda=4$ and second-layer bias radius $\widehat\lambda=1=\delta_0$, as prescribed for the second layer in Theorem~\ref{thm:second}. These bias choices are fixed across all configurations. The first-layer radius is an experimental choice; the theorem's universal constant is not evaluated numerically, so the experiment does not test its explicit sufficient-bias threshold. In each trial, we sample and freeze the random layers, then fit affine readouts on the resulting features. We certify margins over each entire ellipsoidal cell using affine bounds for ReLU and outward-rounded arithmetic; a trial counts as successful only when all required certified margins are strictly positive.
For the first-layer criterion, each $\Phi(E_j^-)$ must be separated with positive margin from $\Phi(S^+)$; the separating hyperplane may depend on $j$. We count a two-layer trial as a joint success only when this first-layer criterion holds and a common positive-margin affine separator exists for $F(S^+)$ and $F(S^-)$. The empirical final readout allows an offset, whereas Theorem~\ref{thm:second} provides a separator with zero output offset; the reported final-layer thresholds therefore describe the broader affine-readout criterion and illustrate width trends rather than directly test the theorem's zero-offset guarantee.

We use nonuniform candidate-width grids, with finer spacing at smaller widths; the first- and second-layer grids range from $8$ to $1024$ and from $8$ to $4096$, respectively. All settings and candidate widths are fixed before running $200$ random trials per configuration. We use the same seed list across configurations, so the comparisons are paired. Among tested width pairs with at least $190$ joint successes, we select the pair minimizing $n+\widehat n$, breaking ties by smaller $n$. The selected first-layer width can therefore exceed the standalone first-layer $95\%$ threshold. The bootstrap intervals in Figure~\ref{fig:exp-width-complexity} are obtained by resampling the paired seeds and reselecting the width pair.

\subsection{Gaussian width and intrinsic dimension}

At fixed $r=128$, we vary $s$ to change $\omega_*^2$ while keeping the parameters in~\eqref{eq:xor-fixed-geometry} fixed. Table~\ref{tab:xor-widths} reports the smallest tested first-layer width achieving at least $190$ successes out of $200$ trials. The ratio of this threshold to $1+\omega_*^2$ ranges from $6.86$ to $8.00$. This near-proportional trend is consistent with the complexity dependence in~\eqref{eq:first-width}; it does not estimate the theorem's unspecified constant.

\begin{table}[htbp]
	\centering
	\caption{Empirical thresholds at $r=128$. The first-layer column gives the smallest tested width with at least $190/200$ successes. The selected pair minimizes $n+\widehat n$ subject to the joint-success criterion; $\omega_*^2$ is rounded.}
	\label{tab:xor-widths}
	\begin{tabular}{@{}rrrrr@{}}
		\toprule
		$s$ & $\omega_*^2$ & First layer alone & $(n,\widehat n)$ & Joint successes\\
		\midrule
		0.06219 & 0.05 & 8  & $(8,12)$   & $190/200$\\
		0.30414 & 0.50 & 12 & $(16,24)$  & $190/200$\\
		0.43946 & 1.00 & 16 & $(20,40)$  & $190/200$\\
		0.54280 & 1.50 & 20 & $(24,80)$  & $192/200$\\
		0.63007 & 2.00 & 24 & $(40,112)$ & $193/200$\\
		0.70724 & 2.50 & 24 & $(56,160)$ & $193/200$\\
		0.77732 & 3.00 & 28 & $(56,256)$ & $191/200$\\
		\bottomrule
	\end{tabular}
\end{table}

\begin{figure}[!htbp]
	\centering
	\includegraphics[trim=0 30 0 20,clip,width=\linewidth]{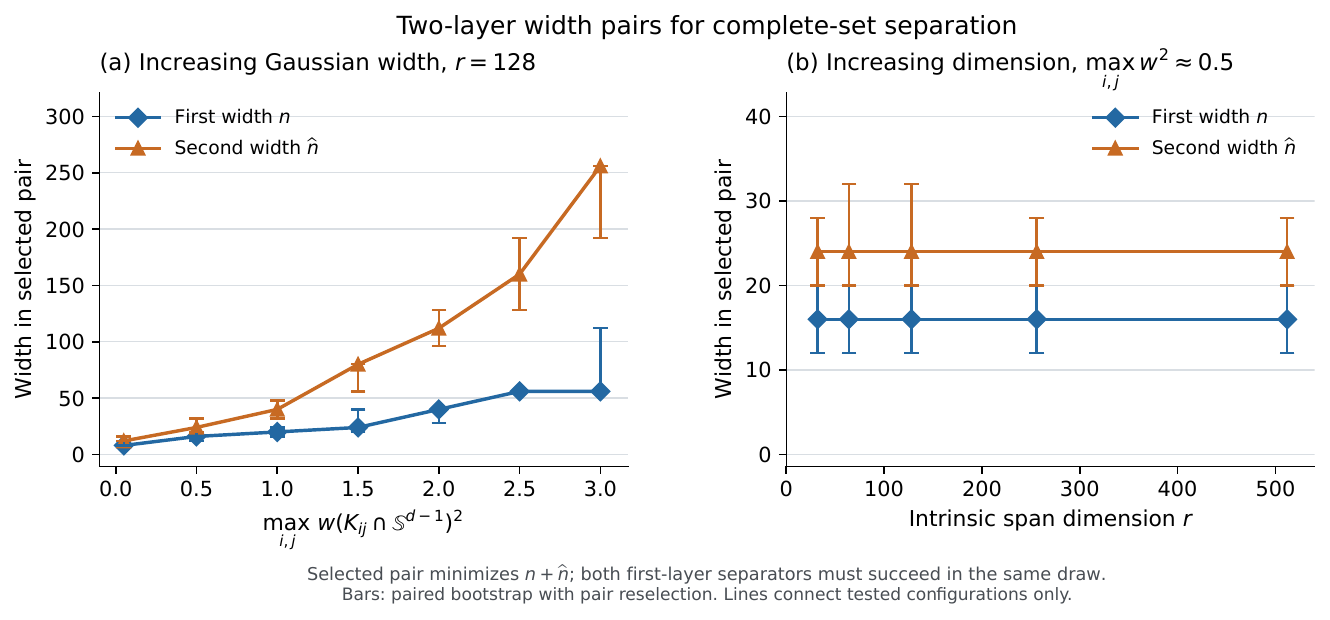}
	\caption{Empirical width pairs based on whole-set separation certificates,
		selected by minimizing $n+\widehat n$ at $95\%$ joint success.
		Blue diamonds show $n$; orange triangles show $\widehat n$.
		(a) Varying $\omega_*^2$ at fixed $r=128$.
		(b) Varying $r$ at nearly fixed $\omega_*^2=0.5$.
		Bars show paired bootstrap $95\%$ intervals with pair reselection.
		Lines connect tested configurations.}
	\label{fig:exp-width-complexity}
\end{figure}

Figure~\ref{fig:exp-width-complexity}(a) and Table~\ref{tab:xor-widths} report the selected width pairs. The selected widths generally increase with $\omega_*^2$, with a more pronounced increase in the second layer. Because $r$ and the other parameters in~\eqref{eq:xor-fixed-geometry} are fixed, this variation is consistent with the Gaussian-width dependence in~\eqref{eq:second-width}.

For Figure~\ref{fig:exp-width-complexity}(b), we take $r=32,64,128,256,512$ and adjust $s$ so that $\omega_*^2\approx0.5$. The standalone first-layer threshold is $12$ for every tested $r$. Each selected pair is $(16,24)$, with $190$--$192$ joint successes out of $200$. Thus, the cone span dimension increases by a factor of $16$, while the selected widths remain unchanged on this grid. As $r$ grows, reducing the transverse semiaxes keeps the Gaussian width nearly constant. These results are consistent with the width bounds' dependence on Gaussian width rather than span dimension alone: for fixed Gaussian width and the other geometric parameters, the sufficient bounds do not increase with $r$.

Taken together, these experiments support Gaussian width as a more
informative measure of separation complexity than span dimension alone.
The term $1+\omega_*^2$ in \eqref{eq:first-width}
and~\eqref{eq:second-width} reflects the effect of set geometry
on the observed network widths.
In particular, high intrinsic dimension need not require wide networks
when Gaussian width and the other parameters remain fixed. 

\FloatBarrier

\section[Conclusions]{Conclusions}
\label{sec:discussion}

This paper establishes geometric conditions for separation by ReLU networks. Positive-margin separation by a finite ReLU network is equivalent to the existence of finite closed convex covers whose opposite-label sets are pairwise at positive distance. Under this condition, two hidden layers always suffice, although one hidden layer may fail. For random networks, pointedness of the difference cones provides a sufficient geometric condition, including for unbounded classes. The first layer separates each $\Phi(E_j^-)$ from $\Phi(S^+)$ and controls all cross-class distances. The second layer combines these cellwise separations into a single positive-margin separator.

The sufficient width of each layer is governed by the squared Gaussian widths of the spherical sections of the difference cones. With the cover sizes, remaining geometric parameters, bias ratios, and failure probability fixed, the width bounds grow at most linearly with the intrinsic span dimension. A linearly inseparable family of translated cones and singletons yields a matching lower bound for each layer. Thus, the quadratic dependence on Gaussian width is optimal uniformly over this family. The experiments use XOR classes that are not linearly separable in the input space. Increasing Gaussian width raises the certified thresholds, although the second-layer thresholds do not exhibit approximately linear growth. At nearly fixed Gaussian width, increasing the intrinsic dimension from $32$ to $512$ leaves both selected widths unchanged.

The counterexamples also clarify the scope of the assumptions. The shallow obstruction shows that two hidden layers are necessary for some classes. The nonpointed-strip example shows that positive distance alone does not guarantee random separation: every fixed finite-depth, finite-width Gaussian ReLU network fails to separate these strips almost surely, even though a deterministic ReLU network separates them. Thus, pointedness cannot simply be omitted from a general random-separation theorem.

\appendix

\section{Proof of the first-layer results}\label{app:first}
This appendix proves the first-layer separation result and provides
the estimates needed for the second layer.
Section~\ref{app:first-margins} derives uniform linear separation margins
from the class gaps and pointedness of the difference cones.
It also bounds the Gaussian width of
$\{(x,\delta_0)/\sqrt{\norm{x}^2+\delta_0^2}:x\in S^+\cup S^-\}$.
Section~\ref{app:refined-first} combines the pairwise separation conditions
using a polynomial and represents it as an expectation of random ReLU features.
Section~\ref{app:first-truncation} uses truncation and error estimates
to obtain a finite-width approximation.
Section~\ref{app:first-features} constructs hyperplanes separating each
$\Phi(E_j^-)$ from $\Phi(S^+)$ and bounds the angular margin
and cross-class distances.
Section~\ref{app:first-norm} bounds $\norm{\Phi(x)}$ and the Gaussian width
of $\{\Phi(x)/\norm{\Phi(x)}:x\in S^+\cup S^-\}$,
and combines the failure probabilities from all steps.

Probabilities in statements about the first layer are taken under the joint distribution of $(W,b)$. Probabilities in statements about the two-layer network are taken under the joint distribution of $(W,b,\widehat W,\widehat b)$. For conditional probabilities in the proofs, we specify which variables are fixed.

\subsection{Separation margins and Gaussian width}\label{app:first-margins}
In this subsection, we derive uniform linear separation margins for
$(x,\delta_0)$ and bound the Gaussian width of
$\{(x,\delta_0)/\sqrt{\norm{x}^2+\delta_0^2}:x\in S^+\cup S^-\}$.
The class gaps control separation at bounded scales, while pointedness
controls separation along unbounded directions.
Together, they yield a margin proportional to
$\sqrt{\norm{x}^2+\delta_0^2}$.
To express these estimates, for any $x\in\R^d$, define
\begin{equation}\label{eq:local-epsilon}
U_x=(x,\delta_0)\in \R^{d+1},\qquad
\varepsilon=\frac{\kappa_{\min}}{16(1+R/\delta_0)^2}.
\end{equation}
When the argument is denoted by $y$, write $U_y=(y,\delta_0)$ accordingly.
\begin{lemma}\label{lem:linear-kappa-lift}
There exists a vector with $\norm{q_{ij}}\le1$ such that
\begin{equation}\label{eq:lifted-sign}
\ip{q_{ij}}{U_x}\ge\varepsilon\norm{U_x}\quad(x\in E_i^+),\qquad
\ip{q_{ij}}{U_y}\le-\varepsilon\norm{U_y}\quad(y\in E_j^-).
\end{equation}
\end{lemma}
\begin{proof}
Apply \eqref{eq:depthineq} from Proposition~\ref{thm:cone}(ii) with $K_{ij}$.
For a pair of sets $E_i^+,E_j^-$, let
\[
p_{ij}=P_{\cl(E_i^+-E_j^-)}(0),\qquad
v_{ij}=p_{ij}/\norm{p_{ij}}.
\]
The projection onto the closed difference set exists, and $\norm{p_{ij}}=\dist(E_i^+,E_j^-)\ge\delta_0$.
The projection inequality gives
\[
\ip{v_{ij}}{x-y}\ge\delta_0
\qquad(x\in E_i^+,\ y\in E_j^-).
\]
To center the threshold for each direction, for $a=u_{ij}$ or $a=v_{ij}$ write
\[
\tau_a=\frac12\left(\inf_{E_i^+}\ip ax+
\sup_{E_j^-}\ip ay\right).
\]
Comparison with the two anchors shows that these infima and suprema are finite, and $|\tau_a|\le R$.
For $x\in E_i^+$, the metric direction gives $\ip{v_{ij}}x-\tau_{v_{ij}}\ge\delta_0/2$;
also, $\ip{u_{ij}}x-\tau_{u_{ij}}>0$, and
$\ip{u_{ij}}x-\tau_{u_{ij}}\ge\kappa_{\min}\norm x-3R$.
Their sum is therefore at least
\[
\max\{\delta_0/2,\ \kappa_{\min}\norm x-3R\}.
\]
The argument for $y\in E_j^-$ is the same. Set
\[
q_{ij}=\frac{\bigl(u_{ij}+v_{ij},
 -(\tau_{u_{ij}}+\tau_{v_{ij}})/\delta_0\bigr)}
 {2\sqrt{1+(R/\delta_0)^2}},
\]
so that $\norm{q_{ij}}\le1$.
The lower bound for $\frac{\ip{q_{ij}}{U_x}}{\norm{U_x}}$
attains its minimum when
$\delta_0/2=\kappa_{\min}\norm{x}-3R$. Hence,
\[
\frac{\kappa_{\min}}
{2\sqrt{1+(R/\delta_0)^2}
	\sqrt{4\kappa_{\min}^2+(6R/\delta_0+1)^2}}
\ge
\frac{\kappa_{\min}}{12(1+R/\delta_0)^2}.
\]
The last inequality uses $0<\kappa_{\min}\le1$ .
Thus \eqref{eq:lifted-sign} holds.

\end{proof}

The subsequent uniform approximation requires control of the complexity of $U_x/\norm{U_x}$. Define the compact set
\[
T_{\delta_0}=\cl\left\{\frac{U_x}{\norm{U_x}}:
 x\in S^+\cup S^-\right\}.
\]
Its closure includes limiting directions of unbounded input sequences. The following width estimate relates its complexity to the difference cones.

Use the anchor $y_1^\circ$ for $S^+$ and $x_1^\circ$ for $S^-$ to decompose each input into an anchor
and a vector in a difference cone. For example, if $x\in E_i^+$, take
$z=(x-y_1^\circ)/\norm{x-y_1^\circ}\in K_{i1}\cap S^{d-1}$. Then
\[
\frac{U_x}{\norm{U_x}}
=\frac{(y_1^\circ,\delta_0)}{\norm{U_x}}
+\frac{\norm{x-y_1^\circ}}{\norm{U_x}}(z,0).
\]
The coefficient satisfies $\norm{x-y_1^\circ}/\norm{U_x}\le1+R/\delta_0$.
For $y\in E_j^-$, using $x_1^\circ$ and $-K_{1j}$ gives the same estimate.
For any nonempty subset $A$ of the unit sphere, fixed $a\in A$,
and standard Gaussian vector $g$ of the same dimension, basic properties of Gaussian width~\cite[Proposition~7.5.2 and Exercise~7.20]{vershynin2018} give
\[
\E_g\sup_{u\in A}|\ip gu|
\le\E_g\sup_{u,u'\in A}\ip g{u-u'}+\E_g|\ip ga|
=2w(A)+\sqrt{2/\pi}.
\]
Taking $g\sim N(0,I_{d+1})$ and applying this comparison to the decomposition above yields
\begin{align*}
w(T_{\delta_0})&\le\E_g\sup_{u\in T_{\delta_0}}|\ip gu|\\
&\le\sqrt{2/\pi}(1+2R/\delta_0)
 +(1+R/\delta_0)\left[2\crosscomplex+\sqrt{2/\pi}\right].
\end{align*}
Hence
\begin{equation}\label{eq:input-complexity}
w(T_{\delta_0})\le C(1+R/\delta_0)\left[1+\crosscomplex\right].
\end{equation}
The constant term in \eqref{eq:input-complexity} comes from the last coordinate $\delta_0$ of $U_x$ and the two anchors.

\subsection{From set margins to random feature representations}\label{app:refined-first}

For each fixed $j \in [N_-]$, we combine the pairwise separation margins into
a single function that distinguishes $E_j^-$ from $S^+$.
We construct this function using a polynomial, and then represent it
as an expectation of Gaussian ReLU features.
By bounding the polynomial degree and the representation coefficients,
we obtain the estimates needed for the finite-width approximation
in the next subsection.
\begin{lemma}[A gap polynomial preserving parity]\label{lem:gap-polynomial}
Let $0<\varepsilon<1$ and $N\ge1$. There exists a polynomial $P$ containing only a linear term and even powers,
of degree $D$ satisfying $c\varepsilon^{-1}\log(4N)\le D\le C\varepsilon^{-1}\log(4N)$, such that
\begin{equation}\label{eq:gap-polynomial-properties}
0\le P(t)\le1,\quad P(t)-P(-t)=t\quad(-1\le t\le1),
\qquad P(-\varepsilon)\le\frac{\varepsilon}{16N}.
\end{equation}
Moreover, $P$ is nondecreasing on $[-1,1]$, and the sum of the absolute values of its nonzero coefficients is at most $e^{CD}$.
\end{lemma}

\begin{proof}
Take the smallest odd integer $L$ at least $4/\varepsilon$, and
$p=\lceil\log_4(16N)\rceil\ge2$.
Let $\mathsf T_L$ denote the Chebyshev polynomial of the first kind. In this proof, define
\begin{gather*}
B(t)=\left[\frac{\mathsf T_L(t)}{Lt}\right]^{2p},\qquad
I=\int_0^1B(s)\,ds,\qquad S(t)=I^{-1}\int_0^tB(s)\,ds,\\
P(t)=\frac12\int_{-1}^t[1+S(s)]\,ds.
\end{gather*}
Use the continuous extension at zero. Since $L$ is odd, $B$ is a nonnegative even polynomial,
and $S$ is an odd, nondecreasing polynomial with $S(\pm1)=\pm1$.
Thus $P'\in[0,1]$ on $[-1,1]$, $P(-1)=0$, $P(1)=1$, and
$P'(t)+P'(-t)=1$.
This proves the first two properties in \eqref{eq:gap-polynomial-properties} and the required parity structure.
The degree is $D=2p(L-1)+2$, and $D$ is comparable to $\varepsilon^{-1}\log(4N)$
up to absolute constants.

The trigonometric identity for Chebyshev polynomials gives
\[
|\mathsf T_L(t)|=|\sin(L\arcsin t)|\ge Lt/2
\]
for $0\le t\le1/(2L)$; also, $|\mathsf T_L(t)|\le1$ for $|t|\le1$.
Hence
\[
I\ge2^{-2p-1}/L,\qquad B(t)\le(Lt)^{-2p}\quad(0<t\le1).
\]
Using $P(-\varepsilon)=(2I)^{-1}
\int_\varepsilon^1(s-\varepsilon)B(s)\,ds$, we obtain
\begin{equation*}
P(-\varepsilon)
\le\frac{2^{2p}\varepsilon}
{(2p-1)(2p-2)(L\varepsilon)^{2p-1}}
\le\varepsilon4^{-p}\le\frac{\varepsilon}{16N}.
\end{equation*}
Finally, the Chebyshev recurrence bounds the sum of the absolute values of the power-basis coefficients of $\mathsf T_L$ by $3^L$.
Submultiplicativity of this coefficient sum under polynomial multiplication bounds the coefficient sum of $B$ by $3^{2pL}$.
The first integration does not increase the coefficient sum, so that of $S$ is at most
\[
I^{-1}3^{2pL}\le L2^{2p+1}3^{2pL}.
\]
Next write $P(t)=P(0)+\int_0^tP'(s)\,ds$. The second integration does not increase the
coefficient sum of the nonconstant terms. Since $0\le P(0)\le1$, the coefficient sum of $P$ is at most
\[
2+L2^{2p+1}3^{2pL}\le e^{CD},
\]
The last step uses $D\asymp pL$ and a sufficiently large positive constant $C$, independent of the sets and $\eta$.

\end{proof}

In Lemma~\ref{lem:gap-polynomial},
take $N=32N_+^2$ and retain $\varepsilon$ from \eqref{eq:local-epsilon}.
The resulting $P$ satisfies $0\le P'\le1$ on $[-1,1]$ and
\begin{equation}\label{eq:refined-polynomial}
P(-\varepsilon)\le\frac{\varepsilon}{512N_+^2},
\qquad
c\varepsilon^{-1}\log(4N_+)\le D
\le C\varepsilon^{-1}\log(4N_+).
\end{equation}
For $x\in\R^d$, use $U_x$ from Lemma~\ref{lem:linear-kappa-lift} to define
\begin{equation}\label{eq:target}
h_j(U_x)=\norm{U_x}\left[
\sum_iP\!\left(\frac{\ip{q_{ij}}{U_x}}{\norm{U_x}}\right)
-\frac\varepsilon2\right].
\end{equation}
Since $\norm{q_{ij}}\le1$, all arguments lie in $[-1,1]$.
For $x\in S^+$, at least one argument is at least $\varepsilon$, and $P(t)\ge t$ for $t\ge0$;
for $y\in E_j^-$, all arguments are at most $-\varepsilon$. Thus
\begin{equation}\label{eq:refined-target-sign}
\begin{aligned}
h_j(U_x)&\ge\tfrac{\varepsilon}{2}\norm{U_x}
&& (x\in S^+),\\
-\tfrac{\varepsilon}{2}\norm{U_y}
\le h_j(U_y)&\le
\left(-\tfrac{\varepsilon}{2}+\tfrac{\varepsilon}{512N_+}\right)\norm{U_y}
&& (y\in E_j^-).
\end{aligned}
\end{equation}
For every $x\in\R^d$, we have $0\le h_j(U_x)/\norm{U_x}+\varepsilon/2\le N_+$.

We next express $h_j$ as an expectation of random ReLU features, so that finitely many neurons can approximate it.
For the probabilists' Hermite polynomials, write
\[
\mu_\ell=\E_G\He_\ell(G)\relu(G),\qquad G\sim N(0,1).
\]
Two integrations by parts give
\[
\mu_0=(2\pi)^{-1/2},\quad \mu_1=1/2,\quad
\mu_{2s}=(-1)^{s-1}(2s-3)!!/\sqrt{2\pi}\quad(s\ge1).
\]
Odd coefficients above degree one vanish. Thus the restriction to a linear term and even powers
ensures that every denominator in the following representation is nonzero.
If $P(t)=\sum_{\ell=0}^D p_\ell t^\ell$, then for
$q\in\R^{d+1}$ with $\norm q\le1$, define
\[
\mathcal H_{q,D}(V)
=\sum_{\substack{\ell\in\{0\}\cup[D]\\ \mu_\ell\ne0}}
\frac{p_\ell\norm q^\ell}{\mu_\ell}
\He_\ell(\ip{q/\norm q}V),
\qquad V\sim N(0,I_{d+1}).
\]
For $q=0$, define $\mathcal H_{0,D}(V)=p_0/\mu_0$.  Fix $x\in\R^d$ and $q$. First condition on
$\ip V{U_x/\norm{U_x}}$ and average the generating function over the orthogonal Gaussian component $P_{U_x^\perp}V$.
Then average over $\ip V{U_x/\norm{U_x}}\sim N(0,1)$ to obtain
\begin{equation}\label{eq:first-hermite}
\E_V\mathcal H_{q,D}(V)\relu(\ip V{U_x})
=\norm{U_x}\,P(\ip q{U_x}/\norm{U_x}).
\end{equation}
Approximating this expectation by finite samples requires moment bounds on its coefficients.
For $G\sim N(0,1)$, Gaussian hypercontractivity
\cite[Theorem~5.1]{janson1997} (see also \cite{odonnell2022}) gives
\[
\norm{\He_\ell(G)}_{L^4}
\le 3^{\ell/2}\norm{\He_\ell(G)}_{L^2}
=3^{\ell/2}\sqrt{\ell!}.
\]
Applying these estimates term by term and using the power-basis coefficient sum, after adjusting the positive constant $C$, independent of the sets and $\eta$, gives
\[
\norm{\mathcal H_{q,D}}_{L^4}
\le C e^{CD}3^{D/2}(D+1)^{3/2}\le e^{CD}.
\]
For $x\in S^+\cup S^-$, since $\ip V{U_x}\sim N(0,\norm{U_x}^2)$, we have
\[
\norm{U_x}=\sqrt{\norm x^2+\delta_0^2}
=\sqrt{2\pi}\,\E_V\relu(\ip V{U_x}).
\]
Thus the expectation representation of $\norm{U_x}$ has the constant coefficient $\sqrt{2\pi}$. Let
\[
\mathcal H_j(V)=\sum_i\mathcal H_{q_{ij},D}(V)
-(\varepsilon/2)\sqrt{2\pi}.
\]
Since $D\ge c\log(4N_+)$, the factor $N_+$ from summation can be absorbed into $e^{CD}$.
Hence $\norm{\mathcal H_j}_{L^4}\le e^{CD}$ and
\[
\E_V\mathcal H_j(V)\relu(\ip V{U_x})=h_j(U_x).
\]

\subsection{Truncation and finite-sample approximation}\label{app:first-truncation}
This subsection derives finite-width approximations of $h_j(U_x)$
and $\norm{U_x}$ from the representations in \eqref{eq:first-hermite}.
Truncation and a change of measure adapt these representations to
the network's uniform biases and give bounded coefficients.
We then control the truncation and sampling errors uniformly
relative to $\norm{U_x}$ over both classes.

Write $V=(g,h)$, where $g\sim N(0,I_d)$ and $h\sim N(0,1)$
are independent, and let $\varphi(t)=(2\pi)^{-1/2}e^{-t^2/2}$.
We truncate $h$ and the representation coefficients.
To bound each target's total relative error by
$\varepsilon/(512N_+)$, take

\begin{equation}\label{eq:refined-cutoffs}
\begin{aligned}
K&=\frac{4096N_+}{\varepsilon}
\max\{1,2\pi,\max_j\norm{\mathcal H_j}_{L^4}^2\},\\
c_0&=\max\left\{1,\sqrt{2\log\left[
128\sqrt3\left(\frac{512N_+}{\varepsilon}\right)^2
\max\{1,2\pi,\max_j\norm{\mathcal H_j}_{L^4}^2\}
\right]}\right\}.
\end{aligned}
\end{equation}
The term $2\pi$ ensures $K>\sqrt{2\pi}$, so the constant coefficient
in the representation of $\norm{U_x}$ is not truncated.
Its truncation error therefore comes only from restricting $|h|\le c_0$. 
For $x\in S^+\cup S^-$ and $j\in[N_-]$,
H\"older's inequality and elementary truncation estimates give
\begin{align*}
\E_V|\mathcal H_j(V)|\relu(\ip V{U_x})\one_{\{|h|>c_0\}}
&\le3^{1/4}\norm{\mathcal H_j}_{L^4}\norm{U_x}
\sqrt{\Pp(|h|>c_0)},\\
\E_V|\mathcal H_j(V)|\relu(\ip V{U_x})\one_{\{|\mathcal H_j|>K\}}
&\le\frac{\norm{\mathcal H_j}_{L^4}^2}{\sqrt2 K}\norm{U_x}.
\end{align*}
By the Gaussian tail bound and \eqref{eq:refined-cutoffs}, these two terms are bounded respectively by
\[
\frac18\frac{\varepsilon}{512N_+}\norm{U_x},
\qquad
\frac1{8\sqrt2}\frac{\varepsilon}{512N_+}\norm{U_x}.
\]
The same estimate for truncating the last Gaussian coordinate applies to the expectation representation of $\norm{U_x}$.

Since
$\log(N_+/\varepsilon)\le CD$ and $\max_j\norm{\mathcal H_j}_{L^4}\le e^{CD}$,
\begin{equation}\label{eq:refined-cutoff-size}
K\le e^{CD},\qquad 1\le c_0^2\le CD.
\end{equation}
These thresholds depend on the geometric parameters and deterministic coefficient moments and are fixed before sampling the network.
Since all $\mathcal H_j$ satisfy the same moment bound, the threshold sizes are uniform over $j\in[N_-]$.

Let $\beta\sim\Unif[-c_0,c_0]$ be independent of $g\sim N(0,I_d)$, and set
\[
B_j(g,\beta)=2c_0\varphi(\beta)\mathcal H_j(g,\beta)
\one_{\{|\mathcal H_j(g,\beta)|\le K\}},
\qquad |B_j|\le c_0K.
\]
The density ratio $2c_0\varphi(\beta)$ rewrites the uniform-bias integral as a truncated Gaussian integral.
For all $x\in S^+\cup S^-$,
\[
\E_{g,\beta}\!\left[B_j(g,\beta)\relu(\ip{(g,\beta)}{U_x})\right]
=\E_V\!\left[\mathcal H_j(V)
\one_{\{|h|\le c_0,\ |\mathcal H_j(V)|\le K\}}
\relu(\ip V{U_x})\right].
\]
For the expectation representation of $\norm{U_x}$, the same change of measure gives
\[
\E_{g,\beta}\!\left[
2c_0\varphi(\beta)\sqrt{2\pi}\,
\relu(\ip{(g,\beta)}{U_x})\right]
=\sqrt{2\pi}\,\E_V\!\left[
\one_{\{|h|\le c_0\}}\relu(\ip V{U_x})\right].
\]
The corresponding coefficient for the uniform-bias representation satisfies
$2c_0\varphi(\beta)\sqrt{2\pi}\le2c_0\le c_0K$.

\begin{lemma}[An empirical estimate with bounded row coefficients]\label{lem:bounded-process}
Let $v=(g,\beta)$ and $v_a=(g_a,\beta_a)$ ($a\in[M]$) be independent and identically distributed,
where $g\sim N(0,I_d)$ and $\beta\sim\Unif[-c_0,c_0]$ are independent and $c_0\ge1$.
For a deterministic nonempty compact set $T\subset S^d$
and $B_j$ satisfying $|B_j|\le B$ for $j\in[J]$,
with probability at least $1-\xi$,
\begin{multline}\label{eq:bounded-process}
\max_{j\in[J]}\sup_{u\in T}
\left|\frac1M\sum_{a=1}^M B_j(v_a)\relu(\ip{v_a}u)
-\E_v B_j(v)\relu(\ip vu)\right|\\
\le\frac{CB}{\sqrt M}
\left[w(T)+c_0\sqrt{\log\frac{2eJ}\xi}\right].
\end{multline}
The coefficients may depend on the entire random row. The probability is over the joint distribution of $\{v_a\}_{a=1}^M$.
\end{lemma}
\begin{proof}
The case $B=0$ is immediate. Assume $B>0$ and $0<\xi<1$.
Fix $j\in[J]$ and $p\ge2$, and write
\[
Z_j=\sup_{u\in T}\left|
\frac1M\sum_a B_j(v_a)\relu(\ip{v_a}u)
-\E_v B_j(v)\relu(\ip vu)\right|.
\]
Let $\epsilon_a$ be independent Rademacher signs, independent of the rows.
For fixed rows, $s\mapsto B_j(v_a)\relu(s)/B$ is $1$-Lipschitz
and vanishes at zero. Symmetrization \cite[Chapter~6]{ledoux1991}
and contraction over the signs \cite[Chapter~4]{ledoux1991},
followed by averaging over the rows, give
\[
\norm{Z_j}_{L^p}\le\frac{4B}{M}
\left\|\sup_{u\in T}\left|
\ip{\sum_a\epsilon_a v_a}u\right|\right\|_{L^p}.
\]

Write $u=(u',u_0)$ and let $P(u',u_0)=u'$, so $PT\subset B_2^d$.
Conditional on $\{\beta_a,\epsilon_a\}_{a=1}^M$,
$\sum_a\epsilon_ag_a\overset{\mathrm d}=\sqrt M\,g$,
where $g\sim N(0,I_d)$.
For a fixed anchor $u_*\in PT$, Gaussian symmetry gives
\begin{align*}
\E_g\sup_{v\in PT}|\ip gv|
&\le\E_g\sup_{v,v'\in PT}\ip g{v-v'}+\E_g|\ip g{u_*}|\\
&\le2w(PT)+\sqrt{2/\pi}.
\end{align*}
Jensen's inequality with an independent $h\sim N(0,1)$ gives
\[
w(PT)\le\E_{g,h}\sup_{u\in T}(\ip g{u'}+hu_0)=w(T).
\]
Since $g\mapsto\sup_{v\in PT}|\ip gv|$ is $1$-Lipschitz,
Gaussian concentration and moment estimates
\cite[Theorem~5.2.3 and Proposition~2.6.6]{vershynin2018} yield
\begin{align*}
\left\|\sup_{u\in T}\left|
\ip{\sum_a\epsilon_a(g_a,0)}u\right|\right\|_{L^p}
&=\sqrt M\left\|\sup_{v\in PT}|\ip gv|\right\|_{L^p}\\
&\le C\sqrt M[w(T)+\sqrt p].
\end{align*}

Conditional on $\{\beta_a\}_{a=1}^M$, the Rademacher moment bound
\cite[Theorem~2.7.5]{vershynin2018} gives
\[
\left(\E_\epsilon\left|\sum_a\epsilon_a\beta_a\right|^p\right)^{1/p}
\le C\sqrt p\left(\sum_a\beta_a^2\right)^{1/2}
\le Cc_0\sqrt{Mp}.
\]
Since $|u_0|\le1$, taking unconditional $L^p$ norms bounds the bias contribution
by $Cc_0\sqrt{Mp}$. The triangle inequality and $c_0\ge1$ therefore give
\[
\norm{Z_j}_{L^p}\le\frac{CB}{\sqrt M}[w(T)+c_0\sqrt p].
\]
Set $p=\max\{2,\log(2eJ/\xi)\}$, so $p\le2\log(2eJ/\xi)$.
Markov's inequality and a union bound yield
\[
\Pp\!\left\{\max_{j\in[J]}Z_j>
\frac{CeB}{\sqrt M}[w(T)+c_0\sqrt p]\right\}
\le Je^{-p}\le\frac{\xi}{2e}\le\xi.
\]
Absorbing the absolute factors into $C$ proves \eqref{eq:bounded-process}.
By continuity, all suprema may be taken over a countable dense subset of $T$,
so the random variables above are measurable.
\end{proof}

Apply Lemma~\ref{lem:bounded-process} simultaneously to the $N_-+1$ targets $h_j(U_x)$ ($j\in[N_-]$) and $\norm{U_x}$, taking
$J=N_-+1$, $B=c_0K$, and $\xi=\eta/24$. The joint empirical error is at most
\[
\frac{Cc_0K}{\sqrt M}\left[
w(T_{\delta_0})+c_0\sqrt{\log\frac{2e(N_-+1)}{\xi}}\right].
\]
Squaring the bracket and absorbing absolute factors inside logarithms into $C$ shows that the error
is at most $\varepsilon/(1024N_+)$ whenever
\begin{equation*}
	M\ge C\left(\frac{N_+}{\varepsilon}\right)^2(c_0K)^2\left[
	w(T_{\delta_0})^2+
	c_0^2\log\frac{24(N_-+1)}{\eta}
	\right].
\end{equation*}
By the positive homogeneity of ReLU, this gives an error at each $U_x$ of at most
\[
\varepsilon\norm{U_x}/(1024N_+)\qquad(x\in S^+\cup S^-).
\]

The Gaussian width estimate for finite unions
\cite[Theorem~5.2.3 and Proposition~2.7.6]{vershynin2018} gives
\begin{equation}\label{eq:union-numerator}
	\crosscomplex
	\le\max_{i,j}w(K_{ij}\cap S^{d-1})
	+C\sqrt{\log(N_+N_-)}.
\end{equation}
Substitute \eqref{eq:input-complexity} and \eqref{eq:union-numerator},
expand the square, and use $(1+R/\delta_0)^2\le1/(16\varepsilon)$.
Bounding both logarithmic terms by $\log(48N_+N_-/\eta)$ reduces the sufficient condition to
\begin{equation}\label{eq:refined-accepted}
	M\ge Ce^{CD}
	\left[1+\max_{i,j}w(K_{ij}\cap S^{d-1})^2
	+\log\frac{48N_+N_-}{\eta}\right].
\end{equation}
Indeed, \eqref{eq:refined-cutoff-size} gives $K\le e^{CD}$ and
$c_0^2\le CD$, while \eqref{eq:refined-polynomial} gives
$\log(N_+/\varepsilon)\le CD$. Thus $(N_+/\varepsilon)^2(c_0K)^2$,
the additional factor $1/\varepsilon$, and the remaining fixed powers can all be absorbed into $e^{CD}$,
with absolute comparison constants.

The two truncation errors are at most $1/8$ and $1/(8\sqrt2)$ of the target error,
and the empirical error is at most $1/2$. These fractions sum to less than one, so each target's total relative error
is less than $\varepsilon/(512N_+)$.

\subsection{\texorpdfstring{$\ip{w_j}{\Phi(x)}$, $\ip r{\Phi(x)}$, and radial separation}{Linear functions and radial separation}}\label{app:first-features}

This subsection uses the finite-sample estimates above to construct
$w_j$ and $r$. The linear function $\ip{w_j}{\Phi(x)}$ separates
$\Phi(E_j^-)$ from $\Phi(S^+)$, while $\ip r{\Phi(x)}$ is comparable
to $\norm{U_x}$. Normalizing $w_j$ then gives an angular separation
margin and cross-class distance bounds.

For the first layer in \eqref{eq:network}, let
$W_{a,\cdot}\in\R^d$ denote row $a$ of $W$ and $b_a$ denote
component $a$ of $b$, where $a\in[n]$.
Each $W_{a,\cdot}\sim N(0,I_d)$ and each
$b_a\sim\Unif[-\lambda,\lambda]$,
with all row vectors and bias components mutually independent.

	\begin{proof}[Proof of Theorem~\ref{thm:first} ]\leavevmode
\paragraph{construction and geometric estimates.} 
Let $\lambda_0=c_0\delta_0$. By \eqref{eq:refined-cutoff-size} and
\eqref{eq:refined-polynomial}, choosing the absolute constant in \eqref{eq:first-bias}
appropriately ensures $\lambda\ge\lambda_0$. Among the row indices $a\in[n]$ in the first layer of \eqref{eq:network},
select those satisfying $|b_a|\le\lambda_0$ and define the actual number of accepted rows by
\[
M=\sum_{a=1}^n\one_{\{|b_a|\le\lambda_0\}}.
\]
Here $M$ is the random count after the network is sampled, rather than the total width $n$; the preceding sample-size condition
is a lower bound required of the actual $M$.

Each row is accepted with probability
$\lambda_0/\lambda$, so
$M\sim\mathrm{Binomial}(n,\lambda_0/\lambda)$. Conditioning on all acceptance indicators
$\{\one_{\{|b_a|\le\lambda_0\}}\}_{a\in[n]}$ makes $M$ deterministic.
Under this conditional distribution, the accepted rows remain independent: $W_{a,\cdot}$ is standard Gaussian,
$b_a/\delta_0$ is $\Unif[-c_0,c_0]$, and the two are independent. This is precisely the distribution
required by the empirical estimate above. We apply that estimate to the accepted rows under this conditional distribution.

For the indicators $\{\one_{\{|b_a|\le\lambda_0\}}\}_{a\in[n]}$,the Chernoff lower-tail bound gives
\[
\Pp\!\left(M<\frac{n\lambda_0}{2\lambda}\right)
\le\exp\!\left(-\frac{n\lambda_0}{8\lambda}\right).
\]
Increasing the absolute constant in \eqref{eq:first-width} and using
\eqref{eq:union-numerator} ensures both $n\lambda_0/\lambda\ge8\log(24/\eta)$
and that $n\lambda_0/(2\lambda)$ is at least the right side of \eqref{eq:refined-accepted}.
Thus, when the acceptance count reaches this threshold, $M$ meets the sample-size condition for joint empirical approximation; the count fails with probability
\begin{equation}\label{eq:first-acceptance-failure}
\Pp\!\left\{M<\frac{n\lambda_0}{2\lambda}\right\}
\le\frac\eta{24}.
\end{equation}

Finally, let $\theta=\sqrt{M/n}$. On the count event above, $c_0\ge1$ gives
\[
\theta\ge\sqrt{\frac{\lambda_0}{2\lambda}}
=\sqrt{\frac{c_0\delta_0}{2\lambda}}
\ge\sqrt{\frac{\delta_0}{2\lambda}}.
\]

At each accepted coordinate $a$, let $v_a=(W_{a,\cdot},b_a/\delta_0)$ and set
\[
w_{j,a}=-\frac{B_j(v_a)}{\sqrt{2M}},\qquad
r_a=\frac{2c_0\varphi(b_a/\delta_0)\sqrt{2\pi}}{\sqrt{2M}};
\]
Set all unaccepted coordinates to zero. Fix any acceptance-indicator configuration satisfying $M\ge n\lambda_0/(2\lambda)$,
and apply the approximation estimate from the preceding subsection under the conditional joint distribution of the accepted rows $\{(W_{a,\cdot},b_a)\}$.
The conditional probability that the following inequalities fail to hold simultaneously is at most $\eta/24$:
\begin{equation}\label{eq:linear-approximations}
\begin{aligned}
\left|\ip{w_j}{\Phi(x)}+\theta h_j(U_x)\right|
&\le\frac{\varepsilon\theta}{512N_+}\norm{U_x},\\
\left|\ip r{\Phi(x)}-\theta\norm{U_x}\right|
&\le\frac{\varepsilon\theta}{512N_+}\norm{U_x},
\qquad x\in S^+\cup S^-,
\end{aligned}
\end{equation}
\begin{equation}\label{eq:coefficient-vector-norm}
\norm{w_j},\norm r\le e^{CD}.
\end{equation}
The bounds on $\norm{w_j}$ and $\norm r$ follow directly from the bounded coefficients on accepted coordinates, and each $r_a$ is nonnegative.
Together with \eqref{eq:refined-target-sign}, since $N_+\ge1$ and $\varepsilon\le1/16$, this gives
\begin{gather}
\ip{w_j}{\Phi(x)}\le-\tfrac\varepsilon4\theta\norm{U_x}
\quad(x\in S^+),\qquad
\ip{w_j}{\Phi(y)}\ge\tfrac\varepsilon4\theta\norm{U_y}
\quad(y\in E_j^-),
\label{eq:raw-sign}\\
|\ip{w_j}{\Phi(x)}|\le(N_++1)\theta\norm{U_x},\qquad
\tfrac12\theta\norm{U_x}\le\ip r{\Phi(x)}
\le\tfrac32\theta\norm{U_x}\quad(x\in S^+\cup S^-).
\label{eq:raw-scale}
\end{gather}

Multiply the conditional failure probability above by the indicator that the count reaches its threshold, then average over
the distribution of $\{\one_{\{|b_a|\le\lambda_0\}}\}_{a\in[n]}$.
This gives the following failure probability over the joint distribution of $(W,b)$:
\begin{equation}\label{eq:first-approximation-failure}
\Pp\!\left\{M\ge\frac{n\lambda_0}{2\lambda},\quad
\text{the inequalities in \eqref{eq:linear-approximations} do not all hold}\right\}
\le\frac\eta{24}.
\end{equation}
The norm bound \eqref{eq:coefficient-vector-norm} follows from deterministic coefficient bounds and adds no failure event.

\paragraph{Normalization of $w_j$, angular margin, and cross-class distances.}
To turn a lower bound on $\ip{w_j}{\Phi(y)-\Phi(x)}$ into an angular margin in feature space, first bound $\norm{\Phi(y)-\Phi(x)}$.

Apply the Gaussian matrix deviation bound \cite{jeong2022} to $W$ on
$T=\cl\bigcup_{i,j}(K_{ij}\cap S^{d-1})$,
taking $t=\sqrt{\log(48/\eta)}$.
If $n\ge C[1+\crosscomplex^2+\log(48/\eta)]$,
the norm deviation divided by $\sqrt n$ is at most the fixed number $3/(2\sqrt2)-1>0$.
By \eqref{eq:union-numerator}, \eqref{eq:first-width} implies this width requirement.
For every $x\in S^+$ and $y\in S^-$, the normalized difference vector belongs to $T$.
Applying the $1$-Lipschitz property of ReLU coordinatewise gives
\[
\norm{\Phi(x)-\Phi(y)}
\le\sqrt{2/n}\norm{W(x-y)}\le\tfrac32\norm{x-y}.
\]
The common bias cancels in the first difference bound; the matrix event depends only on $W$ and holds for every allowed $b$.
Since $2e^{-t^2}=\eta/24$, the failure probability of the cross-class distance upper bound under the joint distribution of $(W,b)$ satisfies
\begin{equation}\label{eq:first-distance-failure}
\Pp\!\left\{
\sup_{\substack{x\in S^+\\y\in S^-}}
\frac{\norm{\Phi(x)-\Phi(y)}}{\norm{x-y}}>\frac32
\right\}\le\frac\eta{24}.
\end{equation}

Let $a_j=w_j/\norm{w_j}$. By \eqref{eq:raw-sign} and
\eqref{eq:coefficient-vector-norm}, $-\ip{a_j}{\Phi(x)}$ and
$\ip{a_j}{\Phi(y)}$ are at least
$\varepsilon\theta/(4e^{CD})$ times $\norm{U_x}$ and $\norm{U_y}$, respectively.
Bounding $\ip{a_j}{\Phi(y)}-\ip{a_j}{\Phi(x)}$ and using
\[
\norm{U_x}+\norm{U_y}\ge\norm{x-y}
\ge\tfrac23\norm{\Phi(x)-\Phi(y)},
\]
gives an angular margin of at least $\varepsilon\theta/(6e^{CD})$.
By $\theta\ge\sqrt{\delta_0/(2\lambda)}$,
$\log(1/\varepsilon)\le CD$, and \eqref{eq:refined-polynomial},
we may take
\begin{equation*}
\vartheta=c\sqrt{\frac{\delta_0}{\lambda}}
\exp\!\left\{-\frac{C(1+R/\delta_0)^2\log(4N_+)}{\kappa_{\min}}\right\}>0.
\end{equation*}
For this choice, all $j$, $x\in S^+$, and $y\in E_j^-$ simultaneously satisfy
\[
\ip{a_j}{\Phi(x)}\le-\vartheta\sqrt{\norm{x}^2+\delta_0^2},
\qquad
\ip{a_j}{\Phi(y)}\ge\vartheta\sqrt{\norm{y}^2+\delta_0^2},
\]
and the bound on $\ip{a_j}{\Phi(y)-\Phi(x)}$ above gives \eqref{eq:first-angle}.
For any $y\in S^-$, choose $j$ such that $y\in E_j^-$.
Since $\norm{a_j}=1$, these score bounds also give
\begin{align*}
\norm{\Phi(y)-\Phi(x)}
&\ge\ip{a_j}{\Phi(y)-\Phi(x)}\\
&\ge\vartheta\bigl(\norm{U_y}+\norm{U_x}\bigr)
\ge\vartheta\norm{x-y}.
\end{align*}
Together with the preceding distance upper bound, this proves \eqref{eq:first-bilip}.
Finally, fix $j\in[N_-]$. For any finite nonnegative linear combination
\[
z=\sum_{\ell=1}^m t_\ell\bigl[\Phi(y_\ell)-\Phi(x_\ell)\bigr],
\qquad t_\ell\ge0,\quad x_\ell\in S^+,\quad y_\ell\in E_j^-,
\]
\eqref{eq:first-angle} and the triangle inequality give
\begin{align*}
\ip{a_j}z
&=\sum_{\ell=1}^m t_\ell\ip{a_j}{\Phi(y_\ell)-\Phi(x_\ell)}\\
&\ge\vartheta\sum_{\ell=1}^m t_\ell\norm{\Phi(y_\ell)-\Phi(x_\ell)}
\ge\vartheta\norm z.
\end{align*}
By continuity of the inner product and norm, passing to limits extends this inequality to all
$z\in\cl\cone(\Phi(E_j^-)-\Phi(S^+))$.
If both $z$ and $-z$ belong to this closed cone, then
\[
0=\ip{a_j}z+\ip{a_j}{-z}\ge2\vartheta\norm z.
\]
Since $\vartheta>0$, this forces $z=0$. The closed cone therefore intersects its negative only at $\{0\}$, so it is pointed.
The probability bookkeeping, including the norm and feature-complexity events, is completed at the end of the proof of Proposition~\ref{prop:first-norm}.

\end{proof}

\subsection{\texorpdfstring{Bounds on $\norm{\Phi(x)}$}{Bounds on the norm of Phi(x)}}\label{app:first-norm}
In this subsection, we prove Proposition~\ref{prop:first-norm}, including \eqref{eq:full-norm}, and
\begin{equation}\label{eq:feature-complexity}
	w\!\left(\cl\left\{\frac{\Phi(x)}{\norm{\Phi(x)}}:
	x\in S^+\cup S^-\right\}\right)^2
	\le C(1+R/\delta_0)^2
	\left[1+\crosscomplex^2+\log\frac{24}\eta\right].
\end{equation}

\begin{proof}[Proof of Proposition~\ref{prop:first-norm} and \eqref{eq:feature-complexity}]
The bound on $\norm{\Phi(x)}$ uses the scale $\lambda$. Write
\[
T_\lambda=\cl\left\{
\frac{(x,\lambda)}{\sqrt{\norm x^2+\lambda^2}}:
 x\in S^+\cup S^-\right\}\subset S^d.
\]
First bound $w(T_\lambda)$ in terms of $w(T_{\delta_0})$.
The first-layer bias condition ensures $\lambda\ge\delta_0$, hence
\begin{equation}\label{eq:lambda-normalized-width}
w(T_\lambda)\le w(T_{\delta_0})+\sqrt{\frac2\pi}.
\end{equation}
To prove this, let $g=(g',g_0)\sim N(0,I_{d+1})$.
Since
\[
\norm{U_x}/\sqrt{\norm x^2+\lambda^2}\le1,\qquad
\lambda/\sqrt{\norm x^2+\lambda^2}\le1,
\]
for all $x\in S^+\cup S^-$ we have
\[
\frac{\ip{g'}x+\lambda g_0}{\sqrt{\norm x^2+\lambda^2}}
\le\left(\frac{\ip{g'}x}{\norm{U_x}}\right)_++(g_0)_+.
\]
Projecting $T_{\delta_0}$ onto its first $d$ coordinates does not increase its Gaussian width.
Choose an anchor in this projection, whose norm is at most $1$. Adding the origin increases the width by at most $1/\sqrt{2\pi}$. Thus
\[
\E_{g'}\sup_{x\in S^+\cup S^-}
\left(\frac{\ip{g'}x}{\norm{U_x}}\right)_+
\le w(T_{\delta_0})+\frac1{\sqrt{2\pi}}.
\]
Using $\E_{g_0}(g_0)_+=1/\sqrt{2\pi}$, take the supremum in the preceding inequality and average over $(g',g_0)$ to obtain
\eqref{eq:lambda-normalized-width}. Taking closures does not change the suprema of these continuous Gaussian processes.

To prove \eqref{eq:full-norm}, fix $u=(u',u_0)\in T_\lambda$
and take independent $g\sim N(0,I_d)$ and $\beta\sim\Unif[-1,1]$.
Conditional Jensen inequalities give
\[
\E_{g,\beta}\relu(\ip g{u'}+\beta u_0)
\ge\max\{\norm{u'}/\sqrt{2\pi},|u_0|/4\}
\ge1/(4\sqrt2).
\]

Apply Lemma~\ref{lem:bounded-process} on $T_\lambda$ to the sample rows
$\{(W_{a,\cdot},b_a/\lambda)\}_{a\in[n]}$, with probability over the joint distribution of $(W,b)$;
take coefficient $1$, $c_0=1$, $J=1$, and $\xi=\eta/24$.
If $n\ge C[1+w(T_\lambda)^2+\log(24/\eta)]$,
the empirical-mean error is at most $1/(8\sqrt2)$.

For $x\in S^+\cup S^-$, take $u=(x,\lambda)/\sqrt{\norm x^2+\lambda^2}$, so that
\[
\Phi(x)/\sqrt{\norm x^2+\lambda^2}
=\sqrt{2/n}\relu(Wu'+(b/\lambda)u_0).
\]
Since the root mean square is at least the arithmetic mean, this proves the lower bound in \eqref{eq:full-norm},
with failure probability satisfying
\begin{equation}\label{eq:first-norm-lower-failure}
\Pp\!\left\{\inf_{x\in S^+\cup S^-}
\frac{\norm{\Phi(x)}}{\sqrt{\norm x^2+\lambda^2}}<\frac18
\right\}\le\frac\eta{24}.
\end{equation}

Projection does not increase Gaussian width.
Thus, under the same width condition, the bound in \cite{jeong2022}
with $t=\sqrt{\log(48/\eta)}$ gives
$\sup_{u\in T_\lambda}\norm{Wu'}/\sqrt n\le11/10$.
Together with $\norm{b/\lambda}\le\sqrt n$ and $|u_0|\le1$,
this yields
\[
\frac{\norm{\Phi(x)}}{\sqrt{\norm x^2+\lambda^2}}
\le\sqrt{2/n}\bigl(\norm{Wu'}+\norm{b/\lambda}|u_0|\bigr)\le3.
\]
The matrix event here depends only on $W$, and the bias bound holds for all $b\in[-\lambda,\lambda]^n$.
Thus, under the joint distribution of $(W,b)$, the failure probability of the upper bound on $\norm{\Phi(x)}$ satisfies
\begin{equation}\label{eq:first-norm-upper-failure}
\Pp\!\left\{\sup_{x\in S^+\cup S^-}
\frac{\norm{\Phi(x)}}{\sqrt{\norm x^2+\lambda^2}}>3
\right\}\le\frac\eta{24}.
\end{equation}
Both bounds on $\norm{\Phi(x)}$ retain the scale $\lambda$.

Next prove \eqref{eq:feature-complexity}. Fix $b$ and, for $u=(u',u_0)\in\R^{d+1}$, write
\[
F_W(u)=\sqrt{2/n}\,\relu\bigl(Wu'+(b/\lambda)u_0\bigr).
\]
By the positive homogeneity of ReLU, for $x\in S^+\cup S^-$,
\[
F_W\!\left(\frac{(x,\lambda)}{\sqrt{\norm x^2+\lambda^2}}\right)
=\frac{\Phi(x)}{\sqrt{\norm x^2+\lambda^2}}.
\]
Fix $W,b$, take $h\sim N(0,I_n)$ independent of the network, and compare the centered Gaussian processes indexed by $u\in T_\lambda$:
\[
X_u=\ip h{F_W(u)},\qquad
Y_u=\sqrt{2/n}\,\ip h{Wu'+(b/\lambda)u_0}.
\]
The $1$-Lipschitz property of ReLU gives
$\E_h(X_u-X_v)^2\le\E_h(Y_u-Y_v)^2$ for $u,v\in T_\lambda$.
Since $F_W(0)=0$, both centered Gaussian processes extend to
$T_\lambda\cup\{0\}$ with $X_0=Y_0=0$.
if one index is zero, we likewise have
\[
\E_h(X_u-X_0)^2=\E_hX_u^2
\le\E_hY_u^2=\E_h(Y_u-Y_0)^2.
\]
The increment-variance comparison therefore holds on all of $T_\lambda\cup\{0\}$.
The Sudakov--Fernique comparison thus gives
\[
w\bigl(F_W(T_\lambda\cup\{0\})\bigr)
=\E_h\sup_{u\in T_\lambda\cup\{0\}}X_u
\le\E_h\sup_{u\in T_\lambda\cup\{0\}}Y_u.
\]
Keep $b$ fixed and average the right side over $W$. First fix the auxiliary vector $h$ and randomize only $W$;
then $W^\top h$ is distributed as $\norm h$ times a standard Gaussian vector, and
\[
\bigl(\ip h{Wu'+(b/\lambda)u_0}\bigr)_+
\le\bigl(\ip{W^\top h}{u'}\bigr)_+
+\left|\ip h{b/\lambda}\right|
\qquad(u\in T_\lambda).
\]
Average first over $W$ and apply the projection and anchor estimates above to $T_\lambda$, then average over the auxiliary vector $h$.
Using $\E_h\norm h\le\sqrt n$ and
$\E_h|\ip h{b/\lambda}|\le\sqrt{2/\pi}\sqrt n$ gives
\[
\E_W\E_h\sup_{u\in T_\lambda}
\bigl(\ip h{F_W(u)}\bigr)_+
\le C[1+w(T_\lambda)].
\]
The $1$-Lipschitz property of ReLU also gives
\[
\sup_{u\in T_\lambda}\norm{F_W(u)-F_{\widetilde W}(u)}
\le\sqrt{2/n}\norm{W-\widetilde W}_{\mathrm F}.
\]
Since $\E_h\norm h\le\sqrt n$, we have
\begin{equation*}
	\left|\E_h\sup_{u\in T_\lambda}\bigl(\ip h{F_W(u)}\bigr)_+
	-\E_h\sup_{u\in T_\lambda}\bigl(\ip h{F_{\widetilde W}(u)}\bigr)_+\right|
	\le\sqrt2\norm{W-\widetilde W}_{\mathrm F}.
\end{equation*}

For each fixed $b\in[-\lambda,\lambda]^n$, apply Gaussian concentration
in $W$. Averaging the resulting bound over $b$ gives
\begin{equation}\label{eq:first-image-concentration-failure}
	\Pp\!\left\{
	\E_h\sup_{u\in T_\lambda}\bigl(\ip h{F_W(u)}\bigr)_+
	> C\left[1+w(T_\lambda)+\sqrt{\log(24/\eta)}\right]
	\right\}
	\le\frac\eta{24}.
\end{equation}

The suprema of these processes can be taken over the compact set $T_\lambda$, so unbounded inputs do not affect the argument or measurability.

On the norm lower-bound event controlled by \eqref{eq:first-norm-lower-failure}, continuity implies
$\norm{F_W(u)}\ge1/8$ for every $u\in T_\lambda$.
When $u=(x,\lambda)/\sqrt{\norm x^2+\lambda^2}$ with $x\in S^+\cup S^-$,
$F_W(u)/\norm{F_W(u)}=\Phi(x)/\norm{\Phi(x)}$.
Fix $(W,b)$ satisfying this norm lower bound. Gaussian symmetry gives
\begin{align*}
w\!\left(\cl\left\{\frac{\Phi(x)}{\norm{\Phi(x)}}:
 x\in S^+\cup S^-\right\}\right)
&\le8\E_h\sup_{u\in T_\lambda}|\ip h{F_W(u)}|\\
&\le16\E_h\sup_{u\in T_\lambda}
\bigl(\ip h{F_W(u)}\bigr)_+.
\end{align*}
The last inequality follows by separating positive and negative parts
and using Gaussian symmetry.
On the norm lower-bound event and the concentration event controlled by
\eqref{eq:first-image-concentration-failure},
\eqref{eq:lambda-normalized-width} and \eqref{eq:input-complexity}
give \eqref{eq:feature-complexity}, with no additional failure event.
The required sample size is
$C[1+w(T_\lambda)^2+\log(24/\eta)]$.
These bounds and \eqref{eq:union-numerator} show that
\eqref{eq:first-width} suffices.

\paragraph{Joint probability.}
Each of the six failure bounds
\eqref{eq:first-acceptance-failure},
\eqref{eq:first-approximation-failure},
\eqref{eq:first-distance-failure},
\eqref{eq:first-norm-lower-failure},
\eqref{eq:first-norm-upper-failure}, and
\eqref{eq:first-image-concentration-failure}
is at most $\eta/24$ under the joint law of $(W,b)$.
The approximation bound already averages over
$\{\one_{\{|b_a|\le\lambda_0\}}\}_{a\in[n]}$
and counts failures only when $M$ meets its threshold,
so no further union bound over $M$ is needed.
Since \eqref{eq:feature-complexity} adds no failure event,
a union bound shows that all conclusions of Theorem~\ref{thm:first}
and Proposition~\ref{prop:first-norm}, together with the auxiliary
estimates for the second layer, hold with probability at least $1-\eta/4$. This establishes Theorem ~\ref{thm:first} with
probability at least $ 1 - \eta $  and completes its proof.

\end{proof}

\section{Proof of separation by the second layer}\label{app:second}

This appendix proves Theorem~\ref{thm:second} in three steps.
Subsection~\ref{app:merge-single-log} constructs an even-degree
homogeneous polynomial $Q$ separating $\Phi(S^+)$ from $\Phi(S^-)$.
It also bounds the degree and coefficients of $Q$.
Subsection~\ref{app:second-representation} gives a random ReLU
expectation with the same separating signs and proves the multiplier
estimates.
Subsection~\ref{app:second-approximation} approximates this expectation
uniformly with finitely many neurons.
This yields a common separating hyperplane for $F(S^+)$ and $F(S^-)$,
together with the required width and margin bounds.

\begin{proof}[Proof of Theorem ~\ref{thm:second}]

Fix a first-layer realization $(W,b)$ satisfying the events in
Appendix~\ref{app:first}, whose total failure probability is at most
$\eta/4$, and retain the notation and estimates established there.
All probabilities below are conditional on this realization until
the final probability estimate. The second layer remains independent
and follows its original distribution.

\subsection{\texorpdfstring{From $\ip{w_j}z$ to a common separating polynomial}{From linear functions to a common separating polynomial}}\label{app:merge-single-log}
This subsection constructs an even-degree homogeneous polynomial
that separates $\Phi(S^+)$ from $\Phi(S^-)$.
Throughout, write $z=\Phi(x)$ with $x\in S^+\cup S^-$.
Each $\ip{w_j}z$ is positive on $\Phi(E_j^-)$ and negative on $\Phi(S^+)$.
Define linear forms $s_j$ that are positive on $\Phi(S^+\cup S^-)$ by

\begin{equation*}
s_j(z)=
\left(\frac{\varepsilon}{2}+\frac{\varepsilon}{128N_+}\right)\ip rz
-\ip{w_j}z.
\end{equation*}
Substituting \eqref{eq:target} into \eqref{eq:linear-approximations} gives, for all $x\in S^+\cup S^-$ and $j\in[N_-]$,
\[
\begin{aligned}
&\left|\frac{s_j(\Phi(x))}{\theta\norm{U_x}}
-\sum_iP\!\left(\frac{\ip{q_{ij}}{U_x}}{\norm{U_x}}\right)
-\frac{\varepsilon}{128N_+}\right|\\
&\qquad\le\left(1+\frac\varepsilon2+\frac\varepsilon{128N_+}\right)
\frac\varepsilon{512N_+}
\le\frac{2\varepsilon}{512N_+}.
\end{aligned}
\]
Here the errors come from $w_j$ and $r$, and $\varepsilon\le1/16$.
The sum lies in $[0,N_+]$.
By \eqref{eq:target} and \eqref{eq:refined-target-sign},
it is at least $\varepsilon$ on $S^+$ and at most
$\varepsilon/(512N_+)$ on $E_j^-$.
Combining these bounds with the preceding error estimate gives
\begin{equation}\label{eq:positive-denominator-bounds}
\begin{aligned}
\frac{\varepsilon\theta}{512N_+}\norm{U_x}
\le s_j(\Phi(x))
&\le\left(N_++\frac{3\varepsilon}{256N_+}\right)
\theta\norm{U_x}
&& (x\in S^+\cup S^-),\\
s_j(\Phi(x))&\ge\tfrac34\varepsilon\theta\norm{U_x}
&& (x\in S^+),\\
s_j(\Phi(y))&\le\frac{\varepsilon\theta}{64N_+}\norm{U_y}
&& (y\in E_j^-).
\end{aligned}
\end{equation}
By \eqref{eq:positive-denominator-bounds}, all $s_j$ are positive on
$\Phi(S^+\cup S^-)$.
Together with \eqref{eq:raw-scale}, this gives
\[
\frac{\varepsilon\ip rz}{2s_j(z)}
\le1 \quad(z\in\Phi(S^+)),\qquad
\frac{\varepsilon\ip rz}{2s_j(z)}
\ge16N_+ \quad(z\in\Phi(E_j^-)).
\]
To make the $j$th powered term on $\Phi(E_j^-)$ dominate
the sum on $\Phi(S^+)$, take
\begin{equation}\label{eq:adaptive-power}
p=\left\lceil\frac{\log(4N_-)}{\log(16N_+)}\right\rceil.
\end{equation}
Here $p\ge1$ and $(16N_+)^p\ge4N_-$.
For $p\le\varepsilon^{-1}$, clearing denominators gives a homogeneous
polynomial of degree $N_-p$.
Otherwise, we use Chebyshev polynomials with common denominator $\ip rz$.
This choice depends only on geometric parameters fixed before sampling. 

\paragraph{Case one: $p\le\varepsilon^{-1}$.}
By these ratio bounds,
\[
\sum_j\left(\frac{\varepsilon\ip rz}{2s_j(z)}\right)^p-2N_-
\le-N_-\quad(z\in\Phi(S^+)),
\]
whereas on each $\Phi(E_j^-)$, the $j$th term is at least $4N_-$ and the other terms are nonnegative,
so the expression is at least $2N_-$. Multiplying by the product of the positive denominators preserves the sign and gives
\begin{equation}\label{eq:reciprocal-merge}
Q(z)=
\left(\frac{\varepsilon\ip rz}{2}\right)^p
\sum_j\prod_{h\ne j}s_h(z)^p
-2N_-\prod_js_j(z)^p.
\end{equation}
On $\Phi(S^+\cup S^-)$, the right-hand side is exactly
\[
\left[\prod_hs_h(z)^p\right]
\left[\sum_j\left(\frac{\varepsilon\ip rz}{2s_j(z)}\right)^p-2N_-\right].
\]
Each term consists of $N_-p$ linear factors; when $N_-=1$, the empty product is $1$.
If $N_-p$ is odd, multiply the right-hand side by $\ip rz$ and still denote it by $Q$.
Then $Q$ is homogeneous of even degree $k\ge2$, with
\[
N_-p\le k\le N_-p+1\le2N_-p.
\]
The positive factor $\ip rz$ preserves the separating signs of $Q$.
The second bracket has absolute value at least $N_-$.
Applying \eqref{eq:positive-denominator-bounds} to each factor and,
in the odd case, using 
$\ip r{\Phi(x)}\ge\theta\norm{U_x}/2$, we obtain
\[
|Q(\Phi(x))|\ge
\begin{cases}
N_-\left(\dfrac{\varepsilon\theta\norm{U_x}}{512N_+}\right)^k,
&k=N_-p,\\[6pt]
N_-\left(\dfrac{\varepsilon\theta\norm{U_x}}{512N_+}\right)^{k-1}
\dfrac{\theta\norm{U_x}}2,
&k=N_-p+1.
\end{cases}
\]
Since $N_-\ge1$ and $\varepsilon/(512N_+)\le1/2$, both cases give
\begin{equation}\label{eq:reciprocal-gap}
|Q(\Phi(x))|\ge
\left(\frac{c\varepsilon}{N_+}\right)^k
\theta^k\norm{U_x}^k
\qquad(x\in S^+\cup S^-).
\end{equation}
We may take $c=1/512$. This strict margin also rules out $Q\equiv0$;
since every term is homogeneous of degree $k$, the actual degree of $Q$ is exactly the even integer $k$.
We next bound the coefficients used in the random feature representation. By \eqref{eq:coefficient-vector-norm},
the coefficient vector of $s_j$ satisfies
\[
\left\|\left(\frac\varepsilon2+\frac\varepsilon{128N_+}\right)r-w_j\right\|
\le2e^{CD}.
\]
The norms of $\varepsilon r/2$ in the numerator and of the possible added factor $r$ are both at most $e^{CD}$.
Keeping the products of linear factors intact, the sum of the absolute scalar coefficients in \eqref{eq:reciprocal-merge} is
$N_-+2N_-=3N_-$. Thus there exists a decomposition
\begin{equation}\label{eq:product-coefficient-bound}
Q(z)=\sum_a c_a\prod_{b=1}^k\ip{v_{ab}}z,\qquad
\sum_a|c_a|\prod_b\norm{v_{ab}}\le e^{CDk}.
\end{equation}
where the absorption in the last step is
\[
\sum_a|c_a|\prod_b\norm{v_{ab}}
\le3N_-(2e^{CD})^k\le e^{CDk};
\]
using $k\ge N_-$ and $D\ge1$ and increasing the positive constant $C$, independent of the sets and $\eta$, as needed.
This decomposition need not be expanded into coordinate monomials. Its coefficient bound matches the Wick isometry estimate below
and introduces no ambient dimension factor.

\paragraph{Case two: $p>\varepsilon^{-1}$.}
In this case, divide each $s_j(z)$ by $\ip rz$ instead.
By \eqref{eq:positive-denominator-bounds} and \eqref{eq:raw-scale},\nopagebreak[4]
\[
0<\frac{s_j(z)}{\ip rz}\le3N_+,\qquad
\frac{s_j(z)}{\ip rz}\ge\frac{\varepsilon}{2}
\ (z\in\Phi(S^+)),\qquad
\frac{s_j(z)}{\ip rz}\le\frac{\varepsilon}{32N_+}
\ (z\in\Phi(E_j^-)).
\]
Apply an affine transformation to the ratios so that their values on $\Phi(S^+)$ lie in the interval where Chebyshev polynomials are bounded:
\begin{equation*}
t_j(z)=1+\frac{\varepsilon}{4N_+}
-\frac{s_j(z)}{2N_+\ip rz}.
\end{equation*}
Specifically, the ratio bounds above give
\[
\begin{aligned}
1\ge t_j(z) &\ge -\tfrac12+\frac\varepsilon{4N_+}
&& (z\in\Phi(S^+)),\\
t_j(z)&\ge1+\frac\varepsilon{4N_+}-\frac\varepsilon{64N_+^2}
\ge1+\frac\varepsilon{8N_+}
&& (z\in\Phi(E_j^-)).
\end{aligned}
\]
Thus the values on $\Phi(S^+)$ lie in $[-1,1]$.
Chebyshev polynomials are bounded on $[-1,1]$
and grow exponentially with the degree to the right of $1$; after squaring, the terms corresponding to all $E_j^-$ are nonnegative.
We therefore take
\begin{equation}\label{eq:common-chebyshev-degree}
k=2\left\lceil
\sqrt{\frac{8N_+}{\varepsilon}}\log(4\sqrt{N_-})
\right\rceil
\end{equation}
and set
\begin{equation}\label{eq:common-chebyshev-merge}
Q(z)=(\ip rz)^k
\left[\sum_j\mathsf T_{k/2}(t_j(z))^2-2N_-\right].
\end{equation}
Here $\mathsf T_{k/2}$ is the Chebyshev polynomial of the first kind, with the notation
of Lemma~\ref{lem:gap-polynomial}.
For $0<a\le1$,
\[
\operatorname{arcosh}(1+a)
=\int_0^a\frac{ds}{\sqrt{2s+s^2}}
\ge\int_0^a\frac{ds}{\sqrt{3s}}
=\frac2{\sqrt3}\sqrt a\ge\sqrt a.
\]
Here $0<\varepsilon/(8N_+)\le1$. Using $\mathsf T_{k/2}(t)=\cosh((k/2)\operatorname{arcosh}t)$ ($t\ge1$),
we have on $\Phi(E_j^-)$
\[
\mathsf T_{k/2}(t_j(z))
\ge\frac12\exp\!\left(\frac k2\sqrt{\frac\varepsilon{8N_+}}\right)
\ge\frac12\exp\bigl(\log(4\sqrt{N_-})\bigr)
=2\sqrt{N_-}.
\]
Hence the $j$th squared term is at least $4N_-$.
The other squared terms are nonnegative, and each term is at most $1$ on $\Phi(S^+)$. Thus
\begin{equation}\label{eq:common-chebyshev-gap}
Q(z)\le-N_-(\ip rz)^k\quad(z\in\Phi(S^+)),\qquad
Q(z)\ge2N_-(\ip rz)^k\quad(z\in\Phi(S^-)).
\end{equation}
Although \eqref{eq:common-chebyshev-merge} is expressed through ratios, it extends to a polynomial on all of $\R^n$.
To see how the denominators cancel, write
$\mathsf T_{k/2}(t)=\sum_{\ell=0}^{k/2}c_\ell t^\ell$. When $\ip rz\ne0$,
\[
\begin{aligned}
&(\ip rz)^{k/2}\mathsf T_{k/2}(t_j(z))\\
&\quad=\sum_{\ell=0}^{k/2}c_\ell
\left[\left(1+\frac\varepsilon{4N_+}\right)\ip rz
-\frac{s_j(z)}{2N_+}\right]^\ell
(\ip rz)^{k/2-\ell}.
\end{aligned}
\]
The right-hand side has no denominators and each term is homogeneous of degree $k/2$, so it is also defined when $\ip rz=0$.
Squaring the right-hand side, summing over $j$, and subtracting $2N_-(\ip rz)^k$ gives an expression for $Q$ on all of $\R^n$.
Every term is homogeneous of degree $k$, and \eqref{eq:common-chebyshev-gap} rules out the zero polynomial, so the actual degree is the even integer $k$.

The Chebyshev recurrence gives $\sum_{\ell=0}^{k/2}|c_\ell|\le3^{k/2}$.
The linear forms used in homogenization are $\ip rz$ and
\[
\left(1+\frac{\varepsilon}{4N_+}\right)\ip rz-\frac{s_j(z)}{2N_+},
\]
and their coefficient vectors all have norm at most $Ce^{CD}$.
Squaring increases the sum of absolute scalar coefficients to at most $3^k$. Summing over $j$ and including
$-2N_-(\ip rz)^k$ gives
\[
\begin{aligned}
\sum_a|c_a|\prod_b\norm{v_{ab}}
&\le(N_-3^k+2N_-)(Ce^{CD})^k\\
&\le3N_-(3Ce^{CD})^k\le e^{CDk}.
\end{aligned}
\]
The last step uses
$k\ge c\log(2N_-)$ from \eqref{eq:common-chebyshev-degree} and $D\ge1$, with only an increase in the absolute constant.
Thus both constructions satisfy \eqref{eq:product-coefficient-bound}.

\paragraph{Unified bounds for the degree and relative margin.}
In case one, $p\le\varepsilon^{-1}$ directly controls $k$; to bound $Dk$,
we combine the denominator of $p$ with the logarithm in $D$. By \eqref{eq:adaptive-power},
\[
\begin{aligned}
p\log(4N_+)
&\le\left(1+\frac{\log(4N_-)}{\log(16N_+)}\right)\log(4N_+)\\
&\le\log(4N_+)+\log(4N_-)\le C\log(4N_+N_-).
\end{aligned}
\]
Thus $k\le2N_-p$ and \eqref{eq:refined-polynomial} give
\[
k\le2N_-\varepsilon^{-1},\qquad
Dk\le CN_-\varepsilon^{-1}p\log(4N_+)
\le CN_-\varepsilon^{-1}\log(4N_+N_-).
\]
The condition $p>\varepsilon^{-1}$ in case two forces $N_-$ to be relatively large.
Since $\varepsilon^{-1}\ge16$, we have $p\ge2$.
By the definition of the ceiling, this requires $\log(4N_-)/\log(16N_+)>1$, hence $N_->4N_+$.
Also,
\[
\varepsilon^{-1}<p\le1+\frac{\log(4N_-)}{\log(16N_+)}
\le C\log(2N_-).
\]
Together with \eqref{eq:common-chebyshev-degree}, this gives
\[
\begin{aligned}
k&\le2+C\sqrt{\frac{N_+}{\varepsilon}}\log(2N_-)\\
&\le C\sqrt{N_-}\,\log^{3/2}(2N_-)\le CN_-.
\end{aligned}
\]
The last comparison uses $\sup_{u\ge1}\log^3(2u)/u=54/e^3<4$, with an absolute constant.
Again by \eqref{eq:refined-polynomial},
\[
Dk\le CN_-\varepsilon^{-1}\log(4N_+)
\le CN_-\varepsilon^{-1}\log(4N_+N_-).
\]
Thus both cases give $k\le CN_-\varepsilon^{-1}$ and the preceding bound for $Dk$.
Substituting $\varepsilon$ yields
\begin{equation}\label{eq:degree}
\begin{aligned}
k&\le\frac{CN_-(1+R/\delta_0)^2}{\kappa_{\min}},\\
Dk&\le
\frac{CN_-(1+R/\delta_0)^2}{\kappa_{\min}}\log(4N_+N_-).
\end{aligned}
\end{equation}
The two cases are $p\le\varepsilon^{-1}$ and $p>\varepsilon^{-1}$, with equality included in case one.
When $N_-=1$, we have $p=1$, so case one applies; multiplying the linear polynomial by $\ip rz$ gives $k=2$.
For $N_+=1$, the logarithms reduce to $\log16$ and $\log4$,
so the bounds remain valid.

To apply random feature approximation, we must also express the margin relative to $\norm{\Phi(x)}$ instead of $\norm{U_x}$.
Since $\lambda\ge\delta_0$,
\[
\frac{\norm{U_x}}{\sqrt{\norm x^2+\lambda^2}}
=\sqrt{\frac{\norm x^2+\delta_0^2}{\norm x^2+\lambda^2}}
\ge\frac{\delta_0}{\lambda}.
\]
Combining the upper bound in \eqref{eq:full-norm} with $\theta\ge\sqrt{\delta_0/(2\lambda)}$ gives
\[
\frac{\theta\norm{U_x}}{\norm{\Phi(x)}}
\ge c\sqrt{\frac{\delta_0}{\lambda}}
\frac{\norm{U_x}}{\sqrt{\norm x^2+\lambda^2}}
\ge c\left(\frac{\delta_0}{\lambda}\right)^{3/2}.
\]
Case one already satisfies \eqref{eq:reciprocal-gap}; in case two,
\eqref{eq:common-chebyshev-gap} and \eqref{eq:raw-scale} give
$|Q(\Phi(x))|\ge(\theta\norm{U_x}/2)^k$, so \eqref{eq:reciprocal-gap} holds as well.
Thus in both cases,
\[
\frac{|Q(\Phi(x))|}{\norm{\Phi(x)}^k}
\ge\left(\frac{c\varepsilon}{N_+}\right)^k
\left(\frac{\delta_0}{\lambda}\right)^{3k/2}
\qquad(x\in S^+\cup S^-).
\]
By the degree lower bound in \eqref{eq:refined-polynomial},
\[
\log\frac{N_+}{c\varepsilon}
\le C\left[\log(4N_+)+\varepsilon^{-1}\right]
\le C\varepsilon^{-1}\log(4N_+)\le CD.
\]
Here $c>0$ is the absolute constant above, so these absorptions are independent of the geometric parameters.
It follows that $|Q(\Phi(x))|/\norm{\Phi(x)}^k$ is at least
$e^{-CDk}(\delta_0/\lambda)^{3k/2}$. Together with the opposite signs already obtained on the two classes, we may take
\begin{equation}\label{eq:relative-gap}
\zeta=e^{-CDk}\left(\frac{\delta_0}{\lambda}\right)^{3k/2},
\end{equation}
so that $Q(z)/\norm z^{k-1}$ is at most $-\zeta\norm z$ on $\Phi(S^+)$
and at least $\zeta\norm z$ on $\Phi(S^-)$.

\subsection{Random ReLU representation of the common polynomial}
\label{app:second-representation}
The Hermite identity gives a random ReLU expectation with the same
separating signs.
The multiplier estimate controls the uniform approximation error
and the coefficient norm in \eqref{eq:coef-moment}.
These bounds preserve the signs and yield a margin bound for $\ip v{F(x)}$.

Let $g\sim N(0,I_n)$ and $G\sim N(0,1)$. For even $k\ge2$, let
$\mu_k(t)=\E_G[\He_k(G)\relu(G+t)]
=\varphi(t)\He_{k-2}(-t)$.
Using \eqref{eq:product-coefficient-bound}, set
\[
\mathcal A(g)=\frac1{\mu_k(0)}
\sum_a c_a:\!\prod_{b=1}^k\ip{v_{ab}}g\!:.
\]
Here the colons denote the Wick polynomial of order $k$ associated with $\prod_{b=1}^k\ip{v_{ab}}g$,
namely the orthogonal projection of this product onto the Hermite subspace of order $k$ in Gaussian $L^2$
\cite[Definition~3.1]{janson1997}. Its generating function is
\[
:\!\prod_{b=1}^k\ip{v_{ab}}g\!:
=\left.\partial_{t_1}\cdots\partial_{t_k}
\exp\!\left(\ip g{\sum_b t_{b}v_{ab}}
-\tfrac12\norm{\sum_b t_bv_{ab}}^2\right)\right|_{t=0}.
\]
Fix $z\ne0$ and a scalar bias $b\in\R$. First condition on $\ip g{z/\norm z}$
and average over the orthogonal Gaussian component $P_{z^\perp}g$, then integrate over $\ip g{z/\norm z}\sim N(0,1)$ to obtain
\begin{equation}\label{eq:second-hermite}
\E_g\mathcal A(g)\relu(\ip gz+b)
=\frac{\mu_k(b/\norm z)}{\mu_k(0)}
\frac{Q(z)}{\norm z^{k-1}}.
\end{equation}
Since all linear factors of $Q$ come from $w_1,\ldots,w_{N_-},r$,
$\mathcal A$ also depends only on their span.
Keeping the first-layer realization, $Q$, $z$, and the scalar bias $b$ fixed,
the Wick isometry formula \cite[Chapters~3--4]{janson1997} gives
\[
\E_g\!\left[\left(:\!\prod_{b=1}^k\ip{v_{ab}}g\!:\right)^2\right]
=\sum_\pi\prod_{b=1}^k\ip{v_{ab}}{v_{a,\pi(b)}}
\le k!\prod_{b=1}^k\norm{v_{ab}}^2,
\]
where $\pi$ runs over all permutations of $[k]$. The $L^2$ triangle inequality gives
\[
\norm{\mathcal A}_{L^2}\le
\frac{\sqrt{k!}}{|\mu_k(0)|}
\sum_a|c_a|\prod_b\norm{v_{ab}}.
\]
To verify \eqref{eq:second-hermite}, fix the value of $\ip g{z/\norm z}$,
average the generating function over the orthogonal component $P_{z^\perp}g$, and then take the mixed derivative at $t=0$. This gives
\[
\E_g\!\left[
:\!\prod_{b=1}^k\ip{v_{ab}}g\!:
\,\middle|\,\ip g{z/\norm z}\right]
=\He_k(\ip g{z/\norm z})
\prod_{b=1}^k\ip{v_{ab}}{z/\norm z}.
\]
Combining this with $\relu(\ip gz+b)=\norm z\,\relu(\ip g{z/\norm z}+b/\norm z)$
and averaging over $\ip g{z/\norm z}\sim N(0,1)$ yields \eqref{eq:second-hermite};
two integrations by parts on $(-t,\infty)$ give the expression for $\mu_k(t)$ above.

By \eqref{eq:product-coefficient-bound} and
$\sqrt{k!}/|\mu_k(0)|\le C(k+1)^{3/2}$, we obtain
\begin{equation}\label{eq:coefficient-bound}
\norm{\mathcal A}_{L^2}\le e^{CDk}.
\end{equation}

For the random ReLU expectation in \eqref{eq:second-hermite} to preserve
the classification signs and relative separation margin of $Q$, we must control the multiplier introduced by the bias,
$\mu_k(b/\norm z)/\mu_k(0)$.
Write $k-2=2h$. The explicit Hermite coefficients give
\[
\left|\frac{\He_{2h}(t)}{\He_{2h}(0)}-1\right|
\le\cosh(\sqrt{2h}|t|)-1.
\]
Thus $\mu_k(t)/\mu_k(0)\ge1/2$ when $|t|\le1/(2\sqrt k)$.
For $k=2$, verify this directly using $e^{-t^2/2}$.
This condition restricts only the size of $|b|/\norm z$ and applies to unbounded feature sets.

When approximating \eqref{eq:second-hermite} by finitely many samples, the coefficient $\mathcal A(g)$ is correlated with the ReLU feature in the same network row and may be unbounded.
The following multiplier estimate uses its polynomial structure and the dimension of the subspace on which it depends to give uniform bounds on high moments.

\begin{lemma}[Polynomial multiplier estimate]\label{lem:multiplier}
Let $g\sim N(0,I_n)$ and let $A:\R^n\to\R$ be a polynomial of total degree at most $k$,
with $\norm A_{L^2}\le B$.
Suppose there exists a fixed $s$-dimensional linear subspace $V\subset\R^n$
such that $A(x)=A(P_Vx)$ for every $x\in\R^n$,
where $P_V$ is the orthogonal projection onto $V$.
Let $(g,\beta)$ and $(g_a,\beta_a)$ ($a\in[M]$) be independent and identically distributed,
with $\beta$ independent of $g$ and $|\beta|\le1$ almost surely.
For a nonempty compact set $T\subset\sqrt2 B_2^{n+1}$, let
\[
Z=\sup_{t\in T}\left|
\frac1M\sum_{a=1}^M A(g_a)\relu(\ip{(g_a,\beta_a)}t)
-\E_{g,\beta} A(g)\relu(\ip{(g,\beta)}t)\right|.
\]
Then for $p\ge2$,
\begin{align}
\norm Z_{L^p}
&\le\frac{C2^{k/2}\sqrt{s+1}\,B}{\sqrt M}
p^{k/2}[w(T)+p],\label{eq:multiplier}\\
 B(p-1)^{k/2} 
&\ge \left\|\left(M^{-1}\sum_a  A(g_a)^2\right)^{1/2}\right\|_{L^p}.\label{eq:coef-moment}
\end{align}
There exists a positive constant $C$, independent of the sets and $\eta$, as well as of $k,s$.
\end{lemma}
\begin{proof}
Write $A_a=A(g_a)$ and let $\epsilon_a$ be independent Rademacher signs,
independent of the sample rows. Minkowski's inequality and Gaussian
hypercontractivity \cite[Theorem~5.1]{janson1997} give
\[
\left\|\left(M^{-1}\sum_a A_a^2\right)^{1/2}\right\|_{L^p}
\le\norm A_{L^p}\le B(p-1)^{k/2},
\]
which proves \eqref{eq:coef-moment}.

For fixed sample rows, set $f_a(u)=A_a\relu(u/A_a)$ if $A_a\ne0$
and $f_a=0$ otherwise. These maps vanish at zero and are $1$-Lipschitz.
Symmetrization and conditional contraction \cite[Chapter~4]{ledoux1991},
applied to $A_a\ip{(g_a,\beta_a)}t$ and then averaged over the rows, give
\[
\norm Z_{L^p}\le\frac4M
\left\|\sup_{t\in T}\left|
\ip{\sum_a\epsilon_a A_a(g_a,\beta_a)}t
\right|\right\|_{L^p}.
\]
Adding the origin does not change the absolute suprema, so $0\in T$
is not required.

Decompose $g_a=g_{a,V}+g_{a,V^\perp}$, where $g_{a,V}=P_Vg_a$,
and write $g_V=P_Vg$.
We bound the $(g_{a,V^\perp},0)$ and $(g_{a,V},\beta_a)$ components separately.
Conditional on $\{g_{a,V},\beta_a,\epsilon_a\}_{a=1}^M$,
the vector $\sum_a\epsilon_a A_ag_{a,V^\perp}$ is centered Gaussian
with covariance $(\sum_a A_a^2)I_{V^\perp}$.
For $P(u,h)=(P_{V^\perp}u,0)$, Gaussian concentration and comparison
with an anchor point
\cite[Proposition~2.6.6, Theorem~5.2.3, and Exercise~7.20]{vershynin2018}
bound the conditional $L^p$ norm of the absolute supremum by
$C(\sum_a A_a^2)^{1/2}[w(PT)+\sqrt p]$.
Averaging and using the sum-of-squares estimate above and
$w(PT)\le w(T)$ \cite[Proposition~7.5.2(g)]{vershynin2018} yield
\[
\left\|\sup_{t\in T}\left|
\ip{\sum_a\epsilon_a A_a(g_{a,V^\perp},0)}t
\right|\right\|_{L^p}
\le C\sqrt M\norm A_{L^p}[w(T)+\sqrt p].
\]

The scalar Rademacher moment bound \cite[Theorem~2.7.5]{vershynin2018}
and Minkowski's inequality give, for deterministic vectors $u_a$,
\[
\left(\E_\epsilon\norm{\sum_a\epsilon_a u_a}^p\right)^{1/p}
\le C\sqrt p\left(\sum_a\norm{u_a}^2\right)^{1/2}.
\]
Condition on all sample rows, apply this with $u_a=A_a(g_{a,V},\beta_a)$,
and average over the rows.
H\"older's inequality and hypercontractivity then give
\begin{align*}
\left\|\sup_{t\in T}\left|
\ip{\sum_a\epsilon_a A_a(g_{a,V},\beta_a)}t
\right|\right\|_{L^p}
&\le C\sqrt{pM}\norm{A(g)(g_V,\beta)}_{L^p}\\
&\le C\sqrt{pM}\norm A_{L^{2p}}\norm{(g_V,\beta)}_{L^{2p}}\\
&\le C\sqrt{pM}\,B(2p-1)^{k/2}
(\sqrt{s+1}+C\sqrt p).
\end{align*}
Here we used $T\subset\sqrt2 B_2^{n+1}$ and
$\norm{(g_V,\beta)}_{L^{2p}}\le\sqrt{s+1}+C\sqrt p$,
which follows from $|\beta|\le1$ and Gaussian concentration.

Combining the two component bounds gives
\begin{align*}
\norm Z_{L^p}\le\frac{CB}{\sqrt M}\bigl\{&
(p-1)^{k/2}[w(T)+\sqrt p]\\
&+\sqrt p(2p-1)^{k/2}(\sqrt{s+1}+C\sqrt p)\bigr\}.
\end{align*}
Using $2p-1\le2p$, $p\ge2$, and $w(T)\ge0$ gives
\eqref{eq:multiplier}, with constants independent of $k,s,p$.
All suprema may be taken over a countable dense subset of $T$,
so the random variables above are measurable.
\end{proof}

\subsection{Finite-width approximation and parameter estimates}
\label{app:second-approximation}
This subsection uses a finite-width network to approximate the preceding
expectation. The resulting width and margin estimates complete the proof
of Theorem~\ref{thm:second}.

\paragraph{Finite-width realization and probability estimates.}Fix a first-layer realization $(W,b)$ satisfying the preceding estimates.
Retain the corresponding $\mathcal A$ and degree $k$, and take
$z\in\Phi(S^+\cup S^-)$.
For the second layer in \eqref{eq:network}, set
$\widehat\lambda=\delta_0$.
Let $\widehat W_{a,\cdot}\in\R^n$ denote row $a$ of $\widehat W$
and $\widehat b_a$ component $a$ of $\widehat b$, where $a\in[\widehat n]$.
Then $\widehat W_{a,\cdot}\sim N(0,I_n)$ and
$\widehat b_a\sim\Unif[-\delta_0,\delta_0]$,
with all row vectors and bias components mutually independent.

Let $g\sim N(0,I_n)$, and let the scalar $\widehat b$ follow the uniform bias distribution
$\Unif[-\delta_0,\delta_0]$ conditional on $|\widehat b|\le\lambda/(16\sqrt k)$,
independently of $g$. Write $f_0(z)=Q(z)/\norm z^{k-1}$ and define
\[
f(z)=\E_{g,\widehat b}
\left[\mathcal A(g)\relu(\ip gz+\widehat b)\right]
=f_0(z)\E_{\widehat b}
\left[\frac{\mu_k(\widehat b/\norm z)}{\mu_k(0)}\right],
\]
where the second equality follows from \eqref{eq:second-hermite}, with $z$ and the first-layer realization fixed.
All sums without a specified range in this section run only over accepted coordinates.

\paragraph{Neuron selection and definition of $v$.}
For $a\in[\widehat n]$, select the neurons satisfying
$|\widehat b_a|\le\lambda/(16\sqrt k)$ and denote the actual number accepted by
\[
M=\sum_{a=1}^{\widehat n}
\one_{\{|\widehat b_a|\le\lambda/(16\sqrt k)\}}.
\]
Here $M$ is a random count, distinct from the total width $\widehat n$.
When $M\ge1$ and the denominator below is nonzero, define the coordinates of $v\in\R^{\widehat n}$ by
\[
v_a=\begin{cases}
-\displaystyle\frac{\mathcal A(\widehat W_{a,\cdot})}
{\bigl(\sum_\ell\mathcal A(\widehat W_{\ell,\cdot})^2\bigr)^{1/2}},
& |\widehat b_a|\le\lambda/(16\sqrt k),\\[6pt]
0,&\text{other coordinates}.
\end{cases}
\]
By definition, $\norm v=1$; the count and approximation bounds below ensure both conditions hold. When $M\ge1$, define
\[
f_M(z)=\frac1M\sum_a\mathcal A(\widehat W_{a,\cdot})
\relu(\ip{\widehat W_{a,\cdot}}z+\widehat b_a).
\]

Each neuron has acceptance probability
$\min\{1,\lambda/(16\delta_0\sqrt k)\}\ge1/(16\sqrt k)$, using $\lambda\ge\delta_0$.
Keep the first layer $(W,b)$ fixed and also fix the values of all acceptance indicators
$\{\one_{\{|\widehat b_a|\le\lambda/(16\sqrt k)\}}\}_{a\in[\widehat n]}$.
Under this conditional distribution, the accepted rows remain independent, with
$\widehat W_{a,\cdot}\sim N(0,I_n)$ and $\widehat b_a$ uniform on
$[-\delta_0,\delta_0]\cap[-\lambda/(16\sqrt k),\lambda/(16\sqrt k)]$,
and the weights and biases are independent. Thus the multiplier estimate applies to these coordinates.

\paragraph{Expectation margin and uniform approximation.}
By \eqref{eq:full-norm}, $\norm z\ge\lambda/8$, so
\[
\frac{|\widehat b_a|}{\norm z}
\le\frac{\lambda/(16\sqrt k)}{\lambda/8}
=\frac1{2\sqrt k}.
\]
Thus $\mu_k(\widehat b_a/\norm z)/\mu_k(0)\ge1/2$ simultaneously for all $z$.
By the definition of $f$ and \eqref{eq:relative-gap},
$f(z)$ is at most $-\zeta\norm z/2$ on $\Phi(S^+)$
and at least $\zeta\norm z/2$ on $\Phi(S^-)$.

Continue to fix the first layer $(W,b)$ and acceptance indicators such that $M\ge1$.
In Lemma~\ref{lem:multiplier}, take
$A=\mathcal A$, $\beta=8\widehat b/\lambda$, and
\[
t_z=\left(\frac z{\norm z},\frac{\lambda}{8\norm z}\right),\qquad
T=\cl\{t_z:z\in\Phi(S^+\cup S^-)\}\subset\sqrt2 B_2^{n+1}.
\]
The selection above ensures $|\beta|\le1$, and the positive homogeneity of ReLU gives
\[
\frac{\relu(\ip gz+\widehat b)}{\norm z}
=\relu\!\left(\ip{(g,\beta)}{t_z}\right).
\]
Thus $Z$ in the lemma is
\[
Z=\sup_{z\in\Phi(S^+\cup S^-)}
\frac{|f_M(z)-f(z)|}{\norm z}.
\]
By \eqref{eq:feature-complexity},
\[
w(T)^2\le C(1+R/\delta_0)^2
\left[1+\crosscomplex^2+\log\frac{24}{\eta}\right].
\]
The last coordinate $\frac{\lambda}{8\norm z}$ increases the Gaussian width by at most
$\sqrt{2/\pi}$.
Both the degree estimate after \eqref{eq:reciprocal-merge} and
\eqref{eq:common-chebyshev-degree} give the degree bound
$k\ge c\log(2N_-)$ for the common separating polynomial $Q$. Together with $s\le N_-+1$, this yields
\[
2^k(s+1)\le 2^k(N_-+2)
=\exp\!\bigl(k\log2+\log(N_-+2)\bigr)
\le e^{Ck}.
\]

Take moment order $p=\log(24/\eta)>2$; in this section, $p$ no longer denotes the power used in \eqref{eq:adaptive-power} to combine the ratios.
By \eqref{eq:multiplier}, to ensure $e\norm Z_{L^p}\le\zeta/4$, it suffices that
\[
M\ge C2^k(s+1)\norm{\mathcal A}_{L^2}^{2}\zeta^{-2}
 p^k[w(T)+p]^2.
\]
Substitute \eqref{eq:coefficient-bound}, \eqref{eq:relative-gap}, and the Gaussian width bound above,
and use \eqref{eq:refined-polynomial} to absorb $(1+R/\delta_0)^2$ and $2^k(s+1)$ into $e^{CDk}$. This gives
\begin{equation}\label{eq:second-accepted}
\begin{aligned}
M\ge{}&\exp(CDk)\left(\frac{\lambda}{\delta_0}\right)^{3k}
\left(\log\frac{24}{\eta}\right)^k\\
&\quad\times\left[1+\crosscomplex^2+\log^2\frac{24}{\eta}\right].
\end{aligned}
\end{equation}
Here $(\lambda/\delta_0)^{3k}$ comes from $\zeta^{-2}$.
Let $M_0$ be the ceiling of the deterministic right-hand side. If
\[
\widehat n\min\{1,\lambda/(16\delta_0\sqrt k)\}
\ge\max\{2M_0,8p\},
\]
we keep the first layer $(W,b)$ fixed and apply the Chernoff bound only to the distribution of acceptance indicators induced by the second-layer bias vector $\widehat b$.
This gives
\begin{equation}\label{eq:second-acceptance-failure}
\Pp\!\left\{
M<\frac{\widehat n}{2}
\min\left\{1,\frac{\lambda}{16\delta_0\sqrt k}\right\}
\,\middle|\,W,b
\right\}
\le e^{-p}=\frac{\eta}{24}.
\end{equation}
Outside this event,
\[
M\ge\frac{\widehat n}{2}
\min\{1,\lambda/(16\delta_0\sqrt k)\}\ge M_0,
\qquad
\sqrt{M/\widehat n}\ge ck^{-1/4}.
\]
Thus a total width $\widehat n\ge C\sqrt k(M_0+p)$ suffices.
Fix the first layer $(W,b)$ and any acceptance indicator configuration satisfying this count condition.
Markov's inequality under the conditional joint distribution of the accepted rows gives
\begin{equation}\label{eq:second-approximation-failure}
\Pp\{Z>\zeta/4\}\le e^{-p}=\frac{\eta}{24},
\end{equation}
where $\Pp$ denotes this conditional probability.
Hence $f_M(z)$ preserves the sign of $f(z)$ and has margin at least $\zeta/4$ relative to $\norm z$.
This conditional probability bound is uniform over all $(W,b)$ satisfying the six groups of first-layer events and all qualifying acceptance indicator configurations.

\paragraph{Coefficient bound and separation margin.}
Still fixing the first layer $(W,b)$ and acceptance indicator configuration above, apply \eqref{eq:coef-moment},
\eqref{eq:coefficient-bound}, and the same Markov estimate under the conditional joint distribution of the accepted rows.
Under the same conditioning, this gives
\begin{equation}\label{eq:second-coefficient-failure}
\Pp\!\left\{
\left(\frac1{2M}\sum_a\mathcal A(\widehat W_{a,\cdot})^2\right)^{1/2}
>e^{CDk}\left(\log\frac{24}{\eta}\right)^{k/2}
\right\}
\le\frac{\eta}{24}.
\end{equation}
On the joint success event, $M\ge M_0\ge1$, and the strict separation margin for $f_M$ in the preceding paragraph ensures
$\sum_a\mathcal A(\widehat W_{a,\cdot})^2>0$, so the vector $v$ defined above is valid.
The second-layer normalization gives
\[
\ip v{\widehat\Phi(z)}
=-\sqrt{\frac{M}{\widehat n}}\,
\frac{f_M(z)}
{\bigl((2M)^{-1}\sum_a\mathcal A(\widehat W_{a,\cdot})^2\bigr)^{1/2}}.
\]
$f_M(z)$ is at most $-\zeta\norm z/4$ on $\Phi(S^+)$
and at least $\zeta\norm z/4$ on $\Phi(S^-)$, so the minus sign gives
the sign orientation in \eqref{eq:final-sign}.
Since $\norm z\ge\lambda/8\ge\delta_0/8$, absorbing $k^{-1/4}$ into $e^{-CDk}$ and using the denominator bound gives the deterministic margin
\begin{equation}\label{eq:second-margin-unabsorbed}
c\delta_0 e^{-CDk}
\left(\frac{\delta_0}{\lambda}\right)^{3k/2}
\left(\log\frac{24}{\eta}\right)^{-k/2}.
\end{equation}

\paragraph{Parameter simplification and joint probability.}
In \eqref{eq:second-accepted}, the factor $e^{CDk}$ uses the bound for $Dk$ in \eqref{eq:degree},
while $(\lambda/\delta_0)^{3k}$ and $(\log(24/\eta))^k$ use only its bound for $k$. To simplify the discussion, write
\[
L=\log\left(\frac{4N_+N_-\lambda}{\delta_0}\log\frac2\eta\right).
\]
Since $\lambda/\delta_0\ge1$ and $0<\eta<1$, we have $L\ge\log(4\log2)>1$ and
\[
\log(4N_+N_-)\le CL,\qquad
\log\frac{2\lambda}{\delta_0}\le CL,\qquad
\log\log\frac{24}{\eta}\le CL.
\]
Substituting the respective bounds in \eqref{eq:degree} gives
\[
CDk+3k\log\frac{\lambda}{\delta_0}
+k\log\log\frac{24}{\eta}
\le\frac{CN_-(1+R/\delta_0)^2}{\kappa_{\min}}L.
\]
On the other hand, \eqref{eq:union-numerator} gives
\begin{equation*}
1+\crosscomplex^2+\log^2\frac{24}{\eta}
\le C\left[1+\max_{i,j}w(K_{ij}\cap S^{d-1})^2\right]e^{CL}.
\end{equation*}
Using the bound for $k$ in \eqref{eq:degree} again,
absorb $\sqrt k$, $p=\log(24/\eta)$, and the factor $e^{CL}$ above
into the preceding exponent estimate.
This proves \eqref{eq:second-width}.
Applying the same estimates to \eqref{eq:second-margin-unabsorbed}, we may take
\begin{equation*}
t=c\delta_0\exp\!\left\{
-\frac{CN_-(1+R/\delta_0)^2}{\kappa_{\min}}
\log\!\left(\frac{4N_+N_-\lambda}{\delta_0}\log\frac2\eta\right)
\right\}>0.
\end{equation*}

Finally, the first-layer events have total failure probability at most
$\eta/4$, as shown in Section~\ref{app:first-norm}.
Fix a first layer satisfying these events.
By \eqref{eq:second-acceptance-failure}, the count condition fails
with conditional probability at most $\eta/24$.
For every acceptance indicator configuration meeting this condition,
\eqref{eq:second-approximation-failure} and
\eqref{eq:second-coefficient-failure} each contribute at most $\eta/24$.
Averaging over these configurations and applying a union bound gives
conditional second-layer failure probability at most $3\eta/24$.
Averaging over $(W,b)$ and adding the first-layer failure probability yields
\[
\frac{\eta}{4}+\frac{3\eta}{24}=\frac{9\eta}{24}<\eta.
\]
The positive constant $C$ is independent of the sets and $\eta$,
and $c$ is an absolute constant.
This proves Theorem~\ref{thm:second}.

\end{proof}
\section{Proofs of necessary width and cone geometry}\label{app:necessary}
This appendix proves the width lower bounds and geometric separation results.
Subsection~\ref{app:deterministic} proves the finite convex-cover
characterization (Theorem~\ref{thm:deterministic}).
Subsection~\ref{app:proof-shallow} shows why one hidden layer may fail
(Example~\ref{ex:shallow}).
Subsection~\ref{app:cone} characterizes random affine separation through
pointedness (Proposition~\ref{thm:cone}).
Subsection~\ref{app:sharp} proves the nonpointed counterexample
(Example~\ref{ex:nonpointed-depth}).
Subsection~\ref{app:cone-width} computes the geometric parameters and
Gaussian width of the lower-bound construction (Example~\ref{ex:numerator-lower}).
Finally, Subsection~\ref{app:kernel-collisions} proves the width lower bounds
in Example~\ref{ex:numerator-lower} by constructing points from different
classes with identical features.

\subsection{Finite convex-cover characterization}\label{app:deterministic}
\begin{proof}[Proof of Theorem~\ref{thm:deterministic}]
Assume (i), with scalar output $h\ge m$ on $S^+$ and $h\le-m$ on $S^-$ for some $m>0$.
The finitely many activation patterns of a finite ReLU network
\cite[Definition~3 and Theorem~2.1]{arora2018} give closed polyhedral cells
$Q_\ell$ covering $\R^d$, on each of which $h$ agrees with an affine map $h_\ell$.
With $\alpha=m/2$, define
\[
C_\ell^+=Q_\ell\cap\{h_\ell\ge\alpha\},
\qquad
C_\ell^-=Q_\ell\cap\{h_\ell\le-\alpha\},
\]
and discard empty cells. A global Lipschitz constant $L_h$ gives, for
$x\in C_\ell^+$ and $y\in C_r^-$,
\[
m=2\alpha\le h(x)-h(y)\le L_h\norm{x-y}.
\]
Thus every opposite-label pair has distance at least $m/L_h$, proving (ii).

Assume (ii), and set $\delta_*:=\min_{i,j}\dist(C_i^+,C_j^-)>0$.
The open convex sets $C_i^++(\delta_*/4)\operatorname{int}B_2^d$ and
$C_j^-+(\delta_*/4)\operatorname{int}B_2^d$ are disjoint.
The strong separation theorem \cite[Theorem~11.4]{rockafellar1970}
gives an affine function $\ell_{ij}$ which, after normalization, satisfies
\begin{equation*}
\ell_{ij}(x)\le-\delta_*/4\quad(x\in C_i^+),
\qquad
\ell_{ij}(y)\ge\delta_*/4\quad(y\in C_j^-).
\end{equation*}
The distance need not be attained.
The first hidden layer contains $z_{ij}(x)=\relu(\ell_{ij}(x))$;
the second contains
\[
q_i(x)=\relu\!\left(\frac{\delta_*}{4}-
\sum_{j=1}^{M_-}z_{ij}(x)\right).
\]
For $x\in S^+$, choosing $i_0$ with $x\in C_{i_0}^+$ gives
$q_{i_0}(x)=\delta_*/4$.
For $y\in S^-$, choose $j_0$ with $y\in C_{j_0}^-$.
Then $z_{ij_0}(y)\ge\delta_*/4$ and $q_i(y)=0$ for every $i$.
Consequently,
\[
h_{\mathrm{out}}(x)=\sum_i q_i(x)-\frac{\delta_*}{8}
\]
has opposite signs with a uniform positive margin, proving (iii).
The implication (iii)$\Rightarrow$(i) is immediate.
\end{proof}

\subsection{Necessity of two hidden layers}\label{app:proof-shallow}
\begin{proof}[Proof of Example~\ref{ex:shallow}]
Suppose that $h>0$ on $S^+$ and $h<0$ on $S^-$.
Absorb terms with $v_k=0$ into $c$ and consider the recession function
\[
h_\infty(v)=\lim_{r\to\infty}\frac{h(rv)}r
=\sum_{k=1}^m a_k\relu(\ip{v_k}{v}).
\]
For $s\in\{-1,1\}^3$, let $O_s=\{v\in\R^3:s_rv_r>0,\ r=1,2,3\}$.
The labels imply $\chi(s)h_\infty(v)\ge0$ on $O_s$.
Here $\chi(s)=s_1s_2s_3\in\{-1,1\}$ is the class label of $E_s$
in~\eqref{eq:orthant-cells}.
Using $\relu(t)=(t+|t|)/2$, write
\[
h_\infty(v)=q\cdot v+\frac12\sum_k a_k|\ip{v_k}{v}|,
\qquad q=\frac12\sum_k a_kv_k.
\]
The inequalities at $v\in O_s$ and $-v\in O_{-s}$ give
$\chi(s)(q\cdot v\pm E(v))\ge0$, where
$E(v)=\frac12\sum_k a_k|\ip{v_k}{v}|$.
Hence $\chi(s)q\cdot v\ge0$.
Approaching each coordinate axis through orthants of opposite parity
forces $q=0$. Thus $h_\infty$ is even, and the opposite labels of
$O_s$ and $O_{-s}$ force $h_\infty=0$ on every open orthant.
Continuity gives $h_\infty\equiv0$ on $\R^3$.

Outside the finitely many ridge hyperplanes $\{\ip{v_k}{v}=0\}$,
the active set along $rv$ is eventually fixed.
Since $h_\infty(v)=0$ cancels the coefficient of $r$,
\[
G(v):=\lim_{r\to\infty}h(rv)
=c+\sum_{\ip{v_k}{v}>0}a_kb_k.
\]
For all sufficiently large $r$, $h(rv)=G(v)$.
Choose $v^0=(0,v_2^0,v_3^0)$ with $v_2^0v_3^0\ne0$,
avoiding ridge hyperplanes that do not contain $v_1=0$.
For sufficiently small $\varepsilon>0$, the active sets at
$v^0\pm\varepsilon e_1$ differ only for normals parallel to $e_1$.
Hence the jump
\[
\begin{aligned}
D&:=G(v^0+\varepsilon e_1)-G(v^0-\varepsilon e_1)\\
&=\sum_{\substack{k\in[m],\,v_k\in\R e_1\\
\ip{v_k}{e_1}>0}}a_kb_k
-\sum_{\substack{k\in[m],\,v_k\in\R e_1\\
\ip{v_k}{e_1}<0}}a_kb_k
\end{aligned}
\]
is independent of the signs of $v_2^0$ and $v_3^0$.
If $v_2^0v_3^0>0$, the labels force $D>0$; if $v_2^0v_3^0<0$,
they force $D<0$. This contradiction proves both assertions.
Restricting a putative separator in $\R^d$ to the first three coordinates
proves the final statement.
\end{proof}

\subsection{Cone geometry of affine separation}\label{app:cone}

\begin{proof}[Proof of Proposition~\ref{thm:cone}]\label{app:proof-cone}
For (ii), pointedness implies $0\notin C_K$: a convex combination of unit
vectors in $K$ equal to zero would give a nonzero vector in $K\cap(-K)$.
Compactness of $C_K$ gives $q=P_{C_K}(0)\ne0$.
The projection inequality $\ip{q}{z}\ge\norm{q}^2$ for $z\in C_K$
yields \eqref{eq:depthineq}.
Testing any unit $u$ at $q\in C_K$ proves optimality;
equality in Cauchy--Schwarz gives uniqueness.

Write $K^*=\{v\in\R^d:\ip vz\ge0\text{ for all }z\in K\}$ for the dual cone.
For (i), pointedness and (ii) give $u_K\in\operatorname{int}K^*$.
Choose a compact spherical cap $B\subset\operatorname{int}K^*$ of positive
surface measure and put
\[
\mu=\inf_{v\in B}\inf_{z\in K\cap S^{d-1}}\ip{v}{z}>0.
\]
The Gaussian weight event $g=rv$ with $v\in B$ and $1\le r\le2$
has positive probability. On this event, writing $\delta=\dist(E^+,E^-)$,
\[
\ip{g}{x^+-x^-}\ge\mu\norm{x^+-x^-}\ge\mu\delta
\]
for every cross pair. The finite support values
$m^+(g)=\inf_{E^+}\ip{g}{x}$ and $M^-(g)=\sup_{E^-}\ip{g}{y}$
satisfy $m^+(g)-M^-(g)\ge\mu\delta$.
Fixed anchors $x_0^\pm\in E^\pm$ and $\norm g\le2$ bound both values
uniformly. For $t=\mu\delta/4$, the admissible biases
\[
[t-m^+(g),-t-M^-(g)]
\]
form an interval of length at least $\mu\delta/2$, contained in a fixed
$[-\lambda,\lambda]$ for sufficiently large $\lambda$.
Thus the conditional bias probability is uniformly positive.
Conversely, separation with margin parameter $t>0$ implies
$\ip{g}{z}\ge0$ on $K$, hence $g\in K^*$.
If $K$ is not pointed, it contains a nonzero line, so $K^*$ lies in a
proper orthogonal hyperplane of Gaussian measure zero, regardless of the bias.
\end{proof}

\subsection{Necessity of pointedness}\label{app:sharp}

\begin{proof}[Proof of Example~\ref{ex:nonpointed-depth}]\label{app:proof-nonpointed-depth}
The opposite difference sets $\R\times[1/2,5/2]$ and
$\R\times[-5/2,-1/2]$ give the stated gaps and cones.
Inductively, almost surely, each $F_\ell$ is either constant or has the form
\[
F_\ell(t,s)=t u+s v+c,\qquad u\ne0,
\]
on $\{(t,s):t\ge T_\ell,\ |s|\le2\}$ for some finite $T_\ell$.
The assertion holds for $F_0(t,s)=t e_1+s e_2$, and constant tails persist.
For a nonconstant tail of $F_{\ell-1}$, condition on the first $\ell-1$ layers.
Under this conditioning,
\[
W_\ell u\sim N(0,\norm{u}^2 I_{n_\ell}),
\]
so all coordinates are nonzero almost surely. Set
\[
T_\ell=\max\!\left\{
T_{\ell-1},\,
1+\max_{a\in[n_\ell]}
\frac{2|(W_\ell v)_a|+|(W_\ell c+b_\ell)_a|}
{|(W_\ell u)_a|}
\right\}.
\]
This threshold is finite almost surely. For all $t\ge T_\ell$ and
$|s|\le2$, each preactivation has the sign of the corresponding
coordinate of $W_\ell u$. The next layer is therefore affine there,
with coefficient of $t$ given by
\[
\sqrt{\frac{2}{n_\ell}}\,\relu(W_\ell u).
\]
This coefficient vanishes only if all preactivation slopes are negative,
in which case the output is identically zero there. Positive normalization factors preserve these
properties, and the finite architecture completes the induction
on a probability-one event.

Thus, for all sufficiently large $t$,
\[
F_q(t,0)=\tfrac12 F_q(t,1)+\tfrac12 F_q(t,-1).
\]
Since $(t,\pm1)\in S^+$ and $(t,0)\in S^-$, this proves
\eqref{eq:nonpointed-hull} and rules out strict affine separation.
Finally, $h(x)=|x_2|-3/4$ has the asserted signs and output gaps.
\end{proof}

\subsection{Cone geometry and Gaussian width}
\label{app:cone-width}
For~\eqref{eq:optimal-circular-cone}, every $z\in K_{\alpha,r}$ has $z_1\ge0$.
The anchors are therefore $\pm3e_1/2$ for $E_1^\pm$ and $\mp e_1/2$ for
$E_2^\pm$, giving $R=3/2$. The intervals joining each class's anchors overlap
on $\{te_1:|t|\le1/2\}$. Thus their convex hulls intersect, which rules out
strict affine separation. The opposite-set differences are
\[
\begin{gathered}
E_1^+-E_1^-=3e_1+K_{\alpha,r},\qquad
E_1^+-E_2^-=E_2^+-E_1^-=e_1+K_{\alpha,r},\\
E_2^+-E_2^-=\{-e_1\}.
\end{gathered}
\]
Their distances are respectively $3,1,1,1$, attained by the anchors, so
$\delta_0=1$. For every $a>0$, the inclusion $ae_1+K_{\alpha,r}\subset K_{\alpha,r}$
and the limit $t^{-1}(ae_1+tz)\to z$ for $z\in K_{\alpha,r}$ give
$\cl\cone(ae_1+K_{\alpha,r})=K_{\alpha,r}$.
This proves the difference-cone identities in~\eqref{eq:lower-family-geometry}.
Every direction in the spherical section of $K_{\alpha,r}$ has first coordinate
at least $\cos\alpha$, while two boundary directions with opposite transverse
components have midpoint $\cos\alpha\,e_1$.
Thus $u_K=e_1$ and $\kappa(K_{\alpha,r})=\cos\alpha$.
The ray $K_{22}$ has depth $1$; its spherical section has Gaussian width
$\E_{g_1}(-g_1)=0$ for $g_1\sim N(0,1)$.
Consequently, $\kappa_{\min}=\cos\alpha$, and the largest squared Gaussian width
is exactly that of $K_{\alpha,r}\cap S^{r-1}$.

Write $(g_1,g_\perp)\sim N(0,I_r)$. Parametrizing the spherical section by
$(\cos\theta,\sin\theta\,v)$, with $0\le\theta\le\alpha$ and $v\in S^{r-2}$, gives
\[
w(K_{\alpha,r}\cap S^{r-1})
=\E_{g_1,g_\perp}\max_{0\le\theta\le\alpha}
\bigl(g_1\cos\theta+\norm{g_\perp}\sin\theta\bigr).
\]
Taking $\theta=\alpha$ and using $\E_{g_1}g_1=0$ gives the lower bound.
For the upper bound, use
\[
g_1\cos\theta+\norm{g_\perp}\sin\theta
\le g_1+(-g_1)_+(1-\cos\alpha)
+\norm{g_\perp}\sin\alpha.
\]
Since $\E_{g_1}(-g_1)_+=1/\sqrt{2\pi}$ and $g_\perp$ has the law of $G_{r-1}$,
\[
w(K_{\alpha,r}\cap S^{r-1})
\le\sin\alpha\,\E_{G_{r-1}}\norm{G_{r-1}}
+\frac{1-\cos\alpha}{\sqrt{2\pi}}.
\]
By $1-\cos\alpha\le\sin\alpha$ and
$\E_{G_{r-1}}\norm{G_{r-1}}\ge\E_{g_1}|g_1|=\sqrt{2/\pi}$,
the last term is at most $\frac12\sin\alpha\,\E_{G_{r-1}}\norm{G_{r-1}}$,
proving~\eqref{eq:circular-width-bounds}.
Cauchy--Schwarz gives the upper bound below; bounding the norm from below
by the sum of absolute coordinates divided by $\sqrt{r-1}$ gives the lower bound:
\[
\sqrt{2(r-1)/\pi}\le\E_{G_{r-1}}\norm{G_{r-1}}\le\sqrt{r-1},
\]
which yields the stated width bounds for $\alpha=\pi/4$.

\subsection{Random kernels and collisions between classes}
\label{app:kernel-collisions}
\begin{proof}[Proof of Example~\ref{ex:numerator-lower}]
Let $q=\min\{n,\widehat n\}$. The probability bound is trivial for $q\ge r$,
so assume $q<r$. We show that intersection of a suitable random kernel
with $\operatorname{int}K_{\alpha,r}$ produces a cross-class pair with identical features.

\paragraph{The first layer is narrower: $n\le\widehat n$.}
Here $n=q<r$. If
$\ker W\cap\operatorname{int}K_{\alpha,r}\ne\varnothing$,
choose $v$ in this intersection. For sufficiently large $t$,
$v-3e_1/(2t)$ remains in the cone, so $x=tv\in E_1^+$ and $-x\in E_1^-$.
Since $Wx=W(-x)=0$, their first-layer features and all subsequent outputs coincide.
This also proves the obstruction for first-layer separation alone, since the
condition for $j=1$ includes this pair.

\paragraph{The second layer is narrower: $\widehat n<n$.}
Set $A=\widehat W W$. Almost surely, $\operatorname{row}W$ has dimension
$\min\{n,r\}>\widehat n$. Conditional on such a $W$, the rows of $A$ are
independent nondegenerate Gaussian vectors in this space. Thus $A$ has full row rank,
and its row space contains none of the nonzero rows $w_a^\top$ of $W$, almost surely.
Averaging over $W$ gives
\[
\ker A\not\subseteq w_a^\perp\quad(a\in[n])
\qquad\text{almost surely}.
\]
Fix weights satisfying these conclusions and arbitrary finite biases $b,\widehat b$.
If $\ker A\cap\operatorname{int}K_{\alpha,r}\ne\varnothing$, this relatively open set
cannot lie in finitely many proper hyperplanes. Choose $v$ in it with
$\ip{w_a}v\ne0$ for every $a$, and let $s_a=\operatorname{sign}\ip{w_a}v$.
Full row rank provides $x_0$ satisfying
$Ax_0=-\widehat W\operatorname{diag}(s)b$.
Because $v$ is interior and all biases are finite, for sufficiently large $t$,
the vector $v+(x_0-3e_1/2)/t$ lies in $K_{\alpha,r}$, and
the point $x=tv+x_0$ satisfies
\[
x\in E_1^+,\quad -x\in E_1^-,\quad
\operatorname{sign}(Wx)=s,\quad
|(Wx)_a|>|b_a|\quad(a\in[n]).
\]
The corresponding preactivations have opposite signs in each coordinate, giving
\begin{equation*}
\relu(Wx+b)-\relu(-Wx+b)
=Wx+\operatorname{diag}(s)b.
\end{equation*}
Since $Av=0$ and $Ax_0=-\widehat W\operatorname{diag}(s)b$,
\[
\widehat W[\Phi(x)-\Phi(-x)]
=\sqrt{2/n}\,[Ax+\widehat W\operatorname{diag}(s)b]=0.
\]
The common second-layer bias cancels, so $F(x)=F(-x)$ for all finite biases.

\paragraph{Intersection probability and the width lower bound.}
Take $L=\operatorname{row}W$ in the first case and $L=\operatorname{row}A$ in the second.
The corresponding matrix has rank $q$ and a right-rotation-invariant distribution,
so $L$ is a uniform $q$-dimensional subspace. For any subspace, maximizing the axial
projection over unit vectors gives
\begin{equation*}
L^\perp\cap\operatorname{int}K_{\alpha,r}\ne\varnothing
\quad\Longleftrightarrow\quad
\norm{P_{L^\perp}e_1}>\cos\alpha
\quad\Longleftrightarrow\quad
\norm{P_Le_1}^2<\sin^2\alpha.
\end{equation*}
Successful separation therefore requires $\norm{P_Le_1}^2\ge\sin^2\alpha$.
Here the law of $L$ is induced by $W$ in the first case and by the joint law
of $(W,\widehat W)$ in the second; $W$ is not fixed in the latter.
Rotational invariance and $\operatorname{tr}P_L=q$ give $\E_L P_L=(q/r)I_r$.
Markov's inequality yields
\[
\Pp\{\text{successful separation}\}
\le\Pp\{\norm{P_Le_1}^2\ge\sin^2\alpha\}
\le\frac{q}{r\sin^2\alpha}.
\]
Since $q=\min\{n,\widehat n\}$ and probabilities are at most $1$,
\begin{equation*}
\Pp\{F(S^+)\text{ and }F(S^-)\text{ are strictly linearly separable}\}
\le\min\left\{1,\frac{\min\{n,\widehat n\}}{r\sin^2\alpha}\right\}.
\end{equation*}
By~\eqref{eq:circular-width-bounds} and
$(\E_{G_{r-1}}\norm{G_{r-1}})^2\le r-1<r$,
\[
w(K_{\alpha,r}\cap S^{r-1})^2
\le\tfrac94(r-1)\sin^2\alpha
\le\tfrac94 r\sin^2\alpha,
\]
which gives~\eqref{eq:both-width-necessary}.
For first-layer separation alone, take $L=\operatorname{row}W$ and $q=n$;
when $n\ge r$, the lower bound holds automatically.
\end{proof}


\begin{funding}
This work was supported in part by the National Natural Science Foundation of China under Grant 12271140, and the Fundamental Research Funds for the Central Universities under Grant B260203007.
\end{funding}


\begin{thebibliography}{99}
	
	\bibitem{alain2016}
	G.~Alain and Y.~Bengio,
	\href{https://openreview.net/forum?id=HJ4-rAVtl}{Understanding intermediate layers using linear classifier probes}.
	In \emph{International Conference on Learning Representations, Workshop Track}, 2017.
	
	\bibitem{arora2018}
	R.~Arora, A.~Basu, P.~Mianjy, and A.~Mukherjee,
	\href{https://openreview.net/forum?id=B1J_rgWRW}{Understanding deep neural networks with rectified linear units}.
	In \emph{International Conference on Learning Representations}, 2018.
	
	\bibitem{bandeira2017}
	A.~S.~Bandeira, D.~G.~Mixon, and B.~Recht,
	Compressive classification and the rare eclipse problem.
	In \emph{Compressed sensing and its applications},
	pp.~197--220, Appl. Numer. Harmon. Anal., Birkh\"auser, Cham, 2017.
	\DOI{10.1007/978-3-319-69802-1\_6}
	
	\bibitem{chandrasekaran2012}
	V.~Chandrasekaran, B.~Recht, P.~A.~Parrilo, and A.~S.~Willsky,
	The convex geometry of linear inverse problems.
	\emph{Found. Comput. Math.} \textbf{12} (2012), no.~6, 805--849.
	\DOI{10.1007/s10208-012-9135-7}
	
	\bibitem{cortes1995}
	C.~Cortes and V.~Vapnik, Support-vector networks.
	\emph{Mach. Learn.} \textbf{20} (1995), no.~3, 273--297.
	\DOI{10.1007/BF00994018}
	
	\bibitem{davenport2010}
	M.~A.~Davenport, P.~T.~Boufounos, M.~B.~Wakin, and R.~G.~Baraniuk,
	Signal processing with compressive measurements.
	\emph{IEEE J. Sel. Top. Signal Process.} \textbf{4} (2010), no.~2, 445--460.
	\DOI{10.1109/JSTSP.2009.2039178}
	
	\bibitem{dirksen2022}
	S.~Dirksen, M.~Genzel, L.~Jacques, and A.~Stollenwerk,
	\href{https://jmlr.org/papers/v23/21-1079.html}{The separation capacity of random neural networks}.
	\emph{J. Mach. Learn. Res.} \textbf{23} (2022), no.~309, 1--47.
	
	\bibitem{ghosal2022}
	P.~Ghosal, S.~Mahankali, and Y.~Sun,
	Randomly initialized one-layer neural networks make data linearly separable.
	2023, \arxiv{2205.11716v2}.
	\DOI{10.48550/arXiv.2205.11716}
	
	\bibitem{gordon1988}
	Y.~Gordon, On Milman's inequality and random subspaces which escape
	through a mesh in $\R^n$.
	In \emph{Geometric aspects of functional analysis: Israel seminar 1986--87},
	pp.~84--106, Lecture Notes in Math. 1317, Springer, Berlin, 1988.
	\DOI{10.1007/BFb0081737}
	
	\bibitem{he2023}
	H.~He and W.~J.~Su, A law of data separation in deep learning.
	\emph{Proc. Natl. Acad. Sci. USA} \textbf{120} (2023), no.~36, e2221704120.
	\DOI{10.1073/pnas.2221704120}
	
	\bibitem{he2015}
	K.~He, X.~Zhang, S.~Ren, and J.~Sun,
	Delving deep into rectifiers: surpassing human-level performance
	on ImageNet classification.
	In \emph{Proceedings of the IEEE International Conference
	on Computer Vision}, pp.~1026--1034, IEEE, 2015.
	\DOI{10.1109/ICCV.2015.123}
	
	\bibitem{hsu2021}
	D.~Hsu, C.~H.~Sanford, R.~A.~Servedio, and E.-V.~Vlatakis-Gkaragkounis,
	\href{https://proceedings.mlr.press/v134/hsu21a.html}{On the approximation power of two-layer networks of random ReLUs}.
	In \emph{Proceedings of the 34th Conference on Learning Theory},
	pp.~2423--2461, Proc. Mach. Learn. Res. 134, 2021.
	
	\bibitem{hu2023}
	M.~Hu, J.~H.~Chion, P.~N.~Suganthan, and R.~K.~Katuwal,
	Ensemble deep random vector functional link neural network for regression.
	\emph{IEEE Trans. Syst. Man Cybern. Syst.} \textbf{53} (2023),
	no.~5, 2604--2615.
	\DOI{10.1109/TSMC.2022.3213628}
	
	\bibitem{huang2006}
	G.-B.~Huang, Q.-Y.~Zhu, and C.-K.~Siew,
	Extreme learning machine: theory and applications.
	\emph{Neurocomputing} \textbf{70} (2006), no.~1--3, 489--501.
	\DOI{10.1016/j.neucom.2005.12.126}
	
	\bibitem{janson1997}
	S.~Janson, \emph{Gaussian Hilbert spaces}.
	Cambridge Tracts in Math. 129, Cambridge University Press, Cambridge, 1997.
	\DOI{10.1017/CBO9780511526169}
	
	\bibitem{jeong2022}
	H.~Jeong, X.~Li, Y.~Plan, and \"O.~Y{\i}lmaz,
	Sub-Gaussian matrices on sets: optimal tail dependence and applications.
	\emph{Comm. Pure Appl. Math.} \textbf{75} (2022), no.~8, 1713--1754.
	\DOI{10.1002/cpa.22024}
	
	\bibitem{ledoux1991}
	M.~Ledoux and M.~Talagrand,
	\emph{Probability in Banach spaces: isoperimetry and processes}.
	Springer, Berlin, Heidelberg, 1991.
	\DOI{10.1007/978-3-642-20212-4}
	
	\bibitem{lei2026minimalwidth}
	Y.~Lei, P.~Wang, Y.~Ying, and D.-X.~Zhou,
		\href{https://www.jmlr.org/papers/v27/24-2030.html}{Optimization and generalization of gradient descent for shallow ReLU networks with minimal width}.
		\emph{J. Mach. Learn. Res.} \textbf{27} (2026), no.~34, 1--35.
	
	\bibitem{lu2021smooth}
	J.~Lu, Z.~Shen, H.~Yang, and S.~Zhang,
		Deep network approximation for smooth functions.
		\emph{SIAM J. Math. Anal.} \textbf{53} (2021), no.~5, 5465--5506.
		\DOI{10.1137/20M134695X}
	
	\bibitem{markowska2021}
	U.~Markowska-Kaczmar and M.~Kosturek,
	Extreme learning machine versus classical feedforward network:
	comparison from the usability perspective.
	\emph{Neural Comput. Appl.} \textbf{33} (2021), no.~22, 15121--15144.
	\DOI{10.1007/s00521-021-06402-y}
	
	\bibitem{masarczyk2023}
	W.~Masarczyk, M.~Ostaszewski, E.~Imani, R.~Pascanu, P.~Mi\l o\'{s}, and T.~Trzci\'{n}ski,
	The tunnel effect: building data representations in deep neural networks.
	In \emph{Advances in Neural Information Processing Systems 36},
	pp.~76772--76805, 2023.
	\DOI{10.52202/075280-3355}
	
	\bibitem{needell2024}
	D.~Needell, A.~A.~Nelson, R.~Saab, P.~Salanevich, and O.~Schavemaker,
	Random vector functional link networks for function approximation on manifolds.
	\emph{Front. Appl. Math. Stat.} \textbf{10} (2024), article no.~1284706.
	\DOI{10.3389/fams.2024.1284706}
	
	\bibitem{odonnell2022}
	R.~O'Donnell, R.~A.~Servedio, and L.-Y.~Tan,
	Fooling Gaussian PTFs via local hyperconcentration.
	In \emph{Proceedings of the 52nd Annual ACM SIGACT Symposium
	on Theory of Computing (STOC '20)},
	pp.~1170--1183, Association for Computing Machinery, New York, 2020.
	\DOI{10.1145/3357713.3384281}
	
	\bibitem{rahimi2007}
	A.~Rahimi and B.~Recht,
	\href{https://proceedings.neurips.cc/paper/2007/hash/013a006f03dbc5392effeb8f18fda755-Abstract.html}{Random features for large-scale kernel machines}.
	In \emph{Advances in Neural Information Processing Systems 20},
	pp.~1177--1184, 2007.
	
	\bibitem{rockafellar1970}
	R.~T.~Rockafellar, \emph{Convex analysis}.
	Princeton Math. Ser. 28, Princeton University Press, Princeton, NJ, 1970.
	\DOI{10.1515/9781400873173}
	
	\bibitem{rudi2017}
	A.~Rudi and L.~Rosasco,
	\href{https://proceedings.neurips.cc/paper/2017/hash/61b1fb3f59e28c67f3925f3c79be81a1-Abstract.html}{Generalization properties of learning with random features}.
	In \emph{Ad\-vances in Neural Information Processing Systems 30},
	pp.~3215--3225, 2017.
	
	\bibitem{salanevich2025}
	P.~Salanevich and O.~Schavemaker,
	Efficient uniform approximation using random vector functional link networks.
	In \emph{2023 International Conference on Sampling Theory and Applications
	(SampTA)}, pp.~1--5, IEEE, 2023.
	\DOI{10.1109/SampTA59647.2023.10301394}
	Corrected version (2025): \arxiv{2306.17501v2}.
	
	\bibitem{scholkopf2002}
	B.~Sch\"olkopf and A.~J.~Smola,
	\emph{Learning with kernels: support vector machines, regularization, optimization, and beyond}.
	MIT Press, Cambridge, MA, 2002.
	\DOI{10.7551/mitpress/4175.001.0001}
	
	\bibitem{shen2022widthdepth}
	Z.~Shen, H.~Yang, and S.~Zhang,
		Optimal approximation rate of ReLU networks in terms of width and depth.
		\emph{J. Math. Pures Appl.} \textbf{157} (2022), 101--135.
		\DOI{10.1016/j.matpur.2021.07.009}
	
	\bibitem{vershynin2018}
	R.~Vershynin, \emph{High-dimensional probability: an introduction with applications in data science}.
	Second ed., Cambridge Ser. Stat. Probab. Math. 58,
	Cambridge University Press, Cambridge, 2026.
	\DOI{10.1017/9781009490672}
	
	\bibitem{xu2026}
	A.~S.~Xu, C.~Yaras, P.~Wang, and Q.~Qu,
	\href{https://proceedings.mlr.press/v300/xu26a.html}{Linearly separable features in shallow nonlinear networks: width scales polynomially with intrinsic data dimension}.
	In \emph{Proceedings of the 29th International Conference on Artificial Intelligence and Statistics},
	pp.~172--180, Proc. Mach. Learn. Res. 300, 2026.
	
	\bibitem{zhang2024logloss}
	Z.~Zhang, L.~Shi, and D.-X.~Zhou,
		\href{https://www.jmlr.org/papers/v25/22-0049.html}{Classification with deep neural networks and logistic loss}.
		\emph{J. Mach. Learn. Res.} \textbf{25} (2024), no.~125, 1--117.

\end{thebibliography}
\end{document}